\documentclass[10pt,twocolumn,letterpaper]{article}

\usepackage[accsupp]{axessibility}  

\usepackage[pagenumbers]{wacv} 

\newcommand{\rev}[1]{{\color{black}#1}}

\usepackage{booktabs}
\usepackage{graphicx}
\usepackage{multirow}
\usepackage{array}
\usepackage[table]{xcolor}
\usepackage{graphicx}
\usepackage{float}
\usepackage{amsmath}
\usepackage{amssymb}
\usepackage{booktabs}
\usepackage{comment}
\usepackage{rotating}
\usepackage[ruled,vlined]{algorithm2e}
\usepackage{xcolor}
\newcommand{\algblockblue}[1]{{\colorbox{blue!10}{\strut \textit{#1}}}}

\newcommand{\algblockgreen}[1]{{\colorbox{green!10}{\strut \textit{#1}}}}

\usepackage{pifont}
\newcommand{\cmark}{\ding{51}}
\newcommand{\xmark}{\ding{55}}

\usepackage{amsthm} \newtheorem{proposition}{Proposition}

\definecolor{wacvblue}{rgb}{0.21,0.49,0.74}
\usepackage[pagebackref,breaklinks,colorlinks,allcolors=wacvblue]{hyperref}

\def\wacvPaperID{881} 
\def\confName{WACV}
\def\confYear{2027}

\title{Adversarial Decoys: Misdirecting Attention-Based Defenses in ViT}

\author{
Giulia Marchiori Pietrosanti \qquad
Giulio Rossolini \qquad
Giorgio Buttazzo\\
Department of Excellence in Robotics and AI, Sant'Anna School of Advanced Studies, Italy\\
{\tt\small \{giulia.marchiori, giulio.rossolini, giorgio.buttazzo\}@santannapisa.it}
}

\begin{document}
\maketitle
\begin{abstract}
Vision Transformers (ViTs) remain vulnerable to localized adversarial attacks, e.g., adversarial patches, while recent test-time defenses mitigate them by suppressing image tokens with abnormally high attention scores. These defenses exploit a strong coupling between attention and adversarial effectiveness: adversarial tokens often need to attract substantial attention to influence the prediction. We introduce \emph{adversarial decoys}, image patches optimized to redirect attention, and therefore related defenses, towards selected target tokens. Rather than jointly optimizing misclassifications and defense evasion, our approach decouples the two objectives: the original adversarial region induces the incorrect prediction, while a separate decoy manipulates the attention ranking used by the defense. A layer-wise objective increases target-token attention and promotes these tokens above competing non-target ones. Since the decoy optimization is not tied to a specific attack formulation, the method can be easily integrated with any existing adversarial patch attack. 
Experiments on ImageNet across multiple ViT architectures and adversarial patch attacks show that decoys consistently redirect high attention scores away from the true adversarial region while preserving much of the attack effectiveness against attention-ranking defenses.
These results reveal a fundamental limitation of using attention alone as an indicator of adversarial relevance\footnote{Code: \href{{https://github.com/GiuliaMP/adversarial-decoys}}{\tt github.com/GiuliaMP/adversarial-decoys}.}.
\end{abstract}

\section{Introduction}
Vision Transformers (ViTs) have achieved remarkable performance across different visual recognition tasks~\cite{dosovitskiy2020image,carion2020end,strudel2021segmenter,oquab2024dinov2}.
A key component of these architectures is self-attention, which enables global interactions among image tokens \cite{NIPS2017_transformers}. Despite these advantages, ViTs remain vulnerable to adversarial input manipulations, including perturbations restricted to localized image regions as adversarial patches \cite{brown2017adversarial}. These can alter the model's inference and steer its prediction toward an incorrect class \cite{fu2022patch,lovisotto2022give,silva2025attacking,pietrosanti2025benchmarking}. 

\begin{figure*}[!t]
\centering
\includegraphics[width=0.95\textwidth]{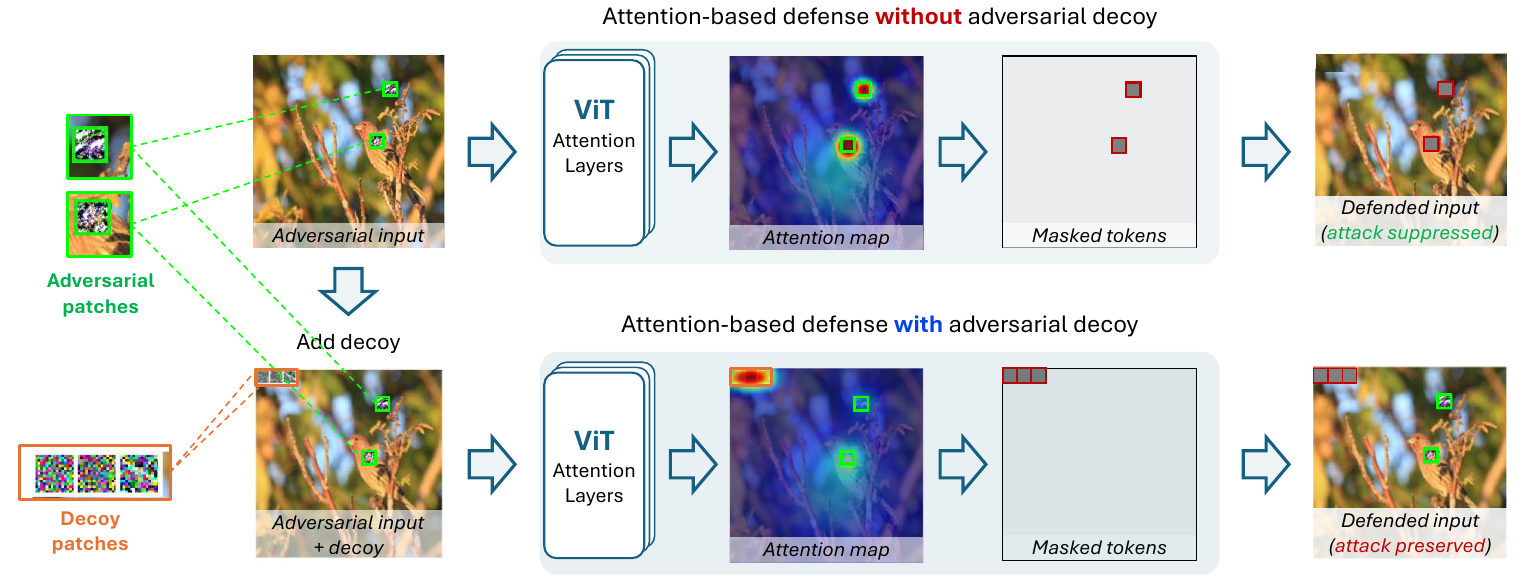}
\vspace{-0.5em}
\caption{\small{\rev{Top: classical adversarial-patch paradigm, where the attention-based defense masks adversarial tokens, suppressing the attack. Bottom: adversarial-decoy paradigm, where optimized decoy patches are added to the already attacked image to redirect the attention-based defense away from the adversarial tokens, preserving attack effectiveness.}}}
\label{fig:intro_idea}
\end{figure*}

Inspired by over-activation analysis in convolutional neural networks \cite{yu2021defending, rossolini2023defending}, recent defenses against localized attacks in transformer-based models exploit the observation that adversarial tokens induce abnormal activation patterns \cite{mu2021defending, liu2023understanding}. Such defenses identify tokens receiving unusually high attention, treat them as suspicious, and mitigate their influence at run-time. Since the attention score of adversarial tokens is closely related to their ability to propagate harmful information, directly reducing it can also weaken the attack, making adaptive evasion particularly challenging (see Sec.~\ref{sec:self_attention} for a more technical discussion).

This raises a central question: \emph{can an attention-based defense be misdirected by introducing highly attended patches, while preserving the effectiveness of the underlying attack?} We address this problem by exploiting the global nature of self-attention, which allows modifications in one region to influence the attention elsewhere in the image. Based on this observation, this work introduces \emph{adversarial decoys}: image patches optimized to attract attention towards attacker-selected target regions. The decoys neither replace the original attack nor are directly optimized to induce misclassifications. Instead, they manipulate the attention distribution so that the selected target tokens dominate the attention used by the defense.

To leverage this, from an evasion perspective, rather than following the conventional adaptive-attack paradigm, in which attacked tokens are jointly optimized both to induce misclassifications and to evade the defense, we decouple the two objectives by combining an adversarial attack with separate decoy patches. The two regions assume complementary roles: the former induces the incorrect prediction, whereas the latter redirects the attention and the related defenses toward specific regions, as illustrated in Fig.~\ref{fig:intro_idea}.

The contributions of the work are the following:
\begin{itemize}
\item \emph{Adversarial decoys} are introduced together with a layer-wise optimization objective that redirects attention toward selected target tokens by (i) increasing their attention across ViT layers and (ii) promoting them above competing non-target tokens in the attention ranking, increasing the likelihood that they are selected by the defense;
\item Optimized decoys are integrated into the attack pipeline as a separate optimization stage and can therefore be combined with different attack strategies for misleading attention-based defenses in ViTs;
\item Experimental analysis is conducted across ViT architectures and different localized attacks on ImageNet \cite{deng2009imagenet}, showing that decoys reduce the overlap between the suppression mask and the true adversarial region, while preserving part of the attack effectiveness.
\end{itemize}
\section{Related work}
\label{sec:RelatedWorks}
\paragraph{Self-attention and robustness in ViTs.} 
Attention is a central mechanism in ViTs and is often treated as a spatial signal for understanding token relevance~\cite{wacv_eliopoulos2025pruning, wacv_patch_ranking}. Prior works have exploited attention patterns for tasks such as object discovery~\cite{simeoni2021localizing, lis2022attentropy} and interpretability~\cite{chefer2021transformer,abnar2020quantifying}. However, attention maps are not always semantically reliable. Recent works \cite{darcet2024vision, jiang2026vision} showed that ViT representations can contain high-norm tokens that do not necessarily correspond to meaningful image content. 
More broadly, while the attention mechanism is informative, it is not always reliable and may be fragile when deliberately manipulated.

Attention has also played a central role in the robustness analysis of ViTs. Early studies observed that ViTs can be more robust than CNNs under some standard adversarial perturbations~\cite{paul2022vision,naseer2021intriguing,aldahdooh2021reveal,benz2021adversarial,shao2022adversarial}, partly due to their weaker reliance on high-frequency features and their global token interactions~\cite{benz2021adversarial,shao2022adversarial}. However, subsequent works showed that ViTs remain highly vulnerable to localized adversarial attacks and transformer-specific perturbations~\cite{bhojanapalli2021understanding,gu2022vision,fu2022patch,lovisotto2022give,silva2025attacking,pietrosanti2025benchmarking}. In the localized setting, such as adversarial patches \cite{brown2017adversarial, adversarial_tshirt, kazoom2026seeing}, the global nature of self-attention can become a security weakness: a manipulated region may influence spatially distant features by attracting their attention, unlike CNNs where the effect of a local perturbation is initially constrained by the receptive field \cite{rossolini2023defending, urtasun_NIPS2016_c8067ad1, yu2021defending}. This makes attention not only a component of ViT robustness, but also an explicit attack surface \cite{silva2025attacking, fu2022patch, lovisotto2022give, alam2025adversarial}. 
For instance, PatchFool~\cite{fu2022patch} augments the classification objective with an attention-aware loss that encourages adversarial tokens to receive high attention. Attention-Fool~\cite{lovisotto2022give} further shows that optimizing post-softmax attention can suffer from saturation, and instead attacks pre-softmax scores to make the adversarial patch attract the attention of all queries.
Alam et al.~\cite{alam2025adversarial} proposed an adversarial attack in the specific context of deformable ViTs, employing additional patches collaboratively to redirect attention towards a target patch containing the adversarial attack. Their collaborative objective is sufficient to strengthen the attack by increasing attention on the target. However, it does not ensure dominance over competing regions, which is instead required in our decoy mechanism to properly steer defenses.

\paragraph{Attention analysis as a defense signal.}
Following the observed relationship between attention of attacked tokens and their adversarial effectiveness, emerging defenses on ViTs use abnormal attention patterns as test-time signals to identify suspicious tokens. A first example is RSA~\cite{mu2021defending}, which detects outlier tokens from their value vectors and neutralizes them during the forward pass rather than masking tokens at the input level. However, RSA is constrained to a single-patch attack setting. Liu et al.~\cite{liu2023understanding} proposed ARMRO, extending this scope to multiple adversarial patches, adopting an explicit token-masking strategy. ARMRO computes a per-token score from post-softmax attention weights across layers and masks tokens selected by a highest-score criterion, significantly improving the robust accuracy of RSA.

These defenses are particularly relevant because they make adaptive attacks based on attention-reduction regularization challenging~\cite{liu2023understanding} (see Sec.~\ref{sec:self_attention} and App.~\ref{app:attention_analysis}): directly reducing the attention of adversarial tokens can also weaken the attack itself, since high attention helps propagate its effect through the network. This motivates a different bypassing paradigm: rather than constraining the attention of adversarial tokens, we investigate whether the defense can be redirected toward a separate region.
\section{Background and motivation}
\label{sec:background}
This section first reviews the standard paradigm underlying localized adversarial attacks against ViTs and the masking-based test-time defenses designed to counter them. Then, it describes the self-attention mechanism of ViTs and provides a mathematical explanation of how state-of-the-art attention-based attacks and defenses exploit its properties.

\subsection{Threat model, attacks and masking defenses}
\label{sec:patch_attacks}
We consider a ViT classifier \( f:\mathcal{X}\rightarrow\mathbb{R}^{Y} \),
where \(\mathcal{X}\subseteq\mathbb{R}^{H\times W\times C}\) is the input space and \(Y\) is the number of classes.
Given an image $x$, the ViT represents it using $n=N+1$ tokens:
\(
[t_0,t_1,\ldots,t_N]\in\mathbb{R}^{n\times d},
\)
where $t_0$ is the \texttt{[CLS]} token and
$\mathcal{I}=\{1,\ldots,N\}$ denotes the set of image-token indices.
A localized adversarial attack modifies only a spatially localized region 
\( \mathcal{A}
\) 
of the input image. Let 
\(
m_{\mathcal{A}}\in\{0,1\}^{H\times W} 
\) 
denote the corresponding pixel-level mask, where \(m_{\mathcal{A},u,v}=1\) if the pixel location \((u,v)\) belongs to \(\mathcal{A}\), and \(0\) otherwise. 
Given adversarial content 
\( p_{\mathcal{A}}\in\mathbb{R}^{H\times W\times C}, 
\) 
the attacked image is built as 
\begin{equation} 
x_{\mathcal{A}} = \left(\mathbf{1}-m_{\mathcal{A}}\right)\odot x + m_{\mathcal{A}}\odot p_{\mathcal{A}}, 
\label{eq:token_replacement} 
\end{equation} 
where the mask is broadcast along the channel dimension.
Given the ground-truth label \(y\), an untargeted localized attack can be formulated as \begin{equation} 
p_{\mathcal{A}}^\star = \arg\max_{p_{\mathcal{A}}\in\mathcal{P}_{\mathcal{A}}} \mathcal{L}_{\mathrm{cls}} \left( f(x_{\mathcal{A}}), y \right), \label{eq:generic_token_attack} 
\end{equation} 
\noindent where \(\mathcal{L}_{\mathrm{cls}}\) is the classification loss and \(\mathcal{P}_{\mathcal{A}}\) denotes the feasible set of adversarial contents restricted to \(\mathcal{A}\).

Conversely, a masking-based defense aims at identifying and suppressing the image regions responsible for the adversarial prediction. 
Let 
\(
m_{\mathrm{def}} \in \{0,1\}^{H \times W}
\)
be a pixel-level mask, denoting the image regions identified by the defense as potentially harmful,
and let \(p_{\mathrm{def}}\) denote the value used to replace the masked pixels, such as zero or the image color mean. The defended image is then built as
\begin{equation}
x_{\mathrm{def}}
=
\left(\mathbf{1} - m_{\mathrm{def}}\right) \odot x_{\mathcal{A}}
+
m_{\mathrm{def}} \odot p_{\mathrm{def}}.
\label{eq:defended_tokens}
\end{equation}
The model then performs a second inference pass on \(x_{\mathrm{def}}\). Note that studies based on ViTs typically apply masking at the token level. In the following, we assume a white-box threat model in which the attacker has full knowledge of the classifier \(f\). The defender is also assumed to have full access to \(f\) during the identification process and masking. 

\subsection{Attention scores in adversarial tokens}
\label{sec:self_attention}
Beyond the general formulation of localized adversarial attacks and masking defenses introduced above, it is useful to recall how attention is computed in ViT architectures and how it can be used to quantify the relevance of individual image tokens.
Let \(X^{l}\in\mathbb{R}^{n\times d}\) denote the token representations entering transformer block 
\(l\in\{1,\ldots,L\}\), where \(X^{1}\) is the initial set of input tokens. 
For each attention head \(h\in\{1,\ldots,N_h\}\), let \(Q^{l,h}\), \(K^{l,h}\), and \(V^{l,h}\) denote the corresponding query, key, and value representations. The attention matrix is defined as 
\begin{equation} 
A^{l,h} = \operatorname{softmax}_{\mathrm{row}} \left( \frac{ Q^{l,h}(K^{l,h})^\top }{ \sqrt{d_k} } \right) \in[0,1]^{n\times n},
\label{eq:self_attention} 
\end{equation} 
\noindent where \(d_k\) is the dimensionality of the query and key representations. 
Each element \(A_{i,j}^{l,h}\) quantifies the attention assigned by query token \(i\) to key token \(j\), and satisfies \( \sum_{j=0}^{n-1} A_{i,j}^{l,h}=1. \) 
To quantify the overall attention received by an image token \(j\), we define its \emph{mean received-attention score} at layer \(l\) as 
\begin{equation} 
s_j^{l} = \frac{1}{N_h|\mathcal{I}|} \sum_{h=1}^{N_h} \sum_{i\in\mathcal{I}} A_{i,j}^{l,h}, \qquad j\in\mathcal{I}. 
\label{eq:received_attention} 
\end{equation} 
This score aggregates the column of the attention matrix associated with token \(j\) by averaging over all attention heads and all image-token queries. The \texttt{[CLS]} token is excluded from both the query set and the set of candidate tokens, so that the analysis focuses exclusively on image tokens.

As discussed in Sec.~\ref{sec:RelatedWorks}, several studies exploit the attention received by adversarial tokens to design both attacks and defenses. 
We formalize this phenomenon in the following and discuss its implications, particularly for the limitations of the design of adaptive attacks against attention-based defenses.

\paragraph{Role of attention in adversarial effect.}
To better understand the relationship between attention and adversarial behavior, we analyze the contribution of a single token to the output of a self-attention head. 
Fixing a layer and an attention head, considering a query token \(i\), let \( o_i = \sum_{j=0}^{n-1} A_{i,j}v_j \) denote the corresponding output, where \(v_j\) is the value representation associated with token \(j\), and \( A_{i,j} = \frac{\exp(z_{i,j})} {\sum_{k=0}^{n-1}\exp(z_{i,k})}, \) with \(z_{i,j}\) denoting the corresponding attention logit. Let \(a\) denote an arbitrary token of interest, and define the renormalized aggregation of all remaining tokens as \( \mu_{i,\neg a} = \sum_{j\neq a} \frac{\exp(z_{i,j})} {\sum_{k\neq a}\exp(z_{i,k})} v_j \), we can thus state 
\begin{proposition} 
\label{prop:attention_token_contribution} 
For a query token \(i\) and an arbitrary token \(a\), the self-attention output satisfies 
\begin{equation} o_i-\mu_{i,\neg a} = A_{i,a} \left( v_a-\mu_{i,\neg a} \right). 
\label{eq:attention_adversarial_effect} 
\end{equation} 
\end{proposition}
A more detailed analysis and derivation of Proposition \ref{prop:attention_token_contribution} are provided in App.~\ref{app:attention_analysis}. 
This proposition highlights the distinct roles played by attention weights and value representations. The term \( (v_a-\mu_{i,\neg a}) \) captures the contribution of token \(a\) in the representation space relative to the remaining tokens, while the attention coefficient \(A_{i,a}\) determines how strongly this contribution is weighted in the output of the query token. When $a$ is an adversarial token, its influence can therefore be increased in two complementary ways: by shifting $(v_a-\mu_{i,\neg a})$ toward a representation-space direction that reduces the classification margin i.e., the adversarial direction, and by increasing $A_{i,a}$, thereby amplifying the effect of this harmful contribution on the attention output.

This explains why attacks such as PatchFool~\cite{fu2022patch} benefit from increasing attention to adversarial tokens and motivates defenses such as ARMRO~\cite{liu2023understanding}, although high attention is not necessarily harmful and adversarial tokens do not need to receive the highest attention.

\section{Adversarial decoy}
As discussed in the previous sections, high attention can provide a signal for detecting adversarial tokens. On the other hand, trying to bypass these defenses by constraining attention while preserving adversarial effectiveness can be challenging.
To this end, rather than constraining the attention associated with the attack, an attacker can further increase the attention directed toward a separate decoy region, thereby redirecting the defense toward innocuous tokens.

In this section, we first introduce the optimization of the decoy patches. We then formalize how the optimized decoy patches can be combined with a localized adversarial attack.

\subsection{Optimization method}
\label{sec:decoy_optimization}
The objective is to modify the predefined decoy patches to induce high received-attention scores within a specific target region. In particular, the optimization aims to make the target tokens dominate the attention ranking across a selected set of transformer layers. Consequently, an attention-based defense may identify these tokens as suspicious and direct its intervention toward them. Formally, let \(\mathcal{I}\) denote the set of image-token indices and let \(\mathcal{D}\subseteq\mathcal{I}\) denote the set of tokens whose spatial patches are associated with the modifiable decoy patches. We further define \(\mathcal{T}\subseteq\mathcal{I}\) as the set of target tokens whose received-attention scores are explicitly optimized, and \({L}_{\mathcal{D}}\) as the set of transformer layers considered during the optimization. In the general formulation, the modifiable region \(\mathcal{D}\) and the target region \(\mathcal{T}\) may overlap or be disjoint. 

For each layer \(l\in{L}_{\mathcal{D}}\), let \(s_j^l\) denote the
received-attention score of image token \(j\), as defined in Eq.~\ref{eq:received_attention}. As a first objective, the optimization
encourages all target tokens to receive high attention. We define their
average received-attention score at layer \(l\) as
\(
    \bar{s}_{\mathcal{T}}^{l}
    =
    \frac{1}{|\mathcal{T}|}
    \sum_{i\in \mathcal{T}}
    s_i^l.
    \label{eq:mean_target_attention}
\)
The corresponding first objective is
\begin{equation}
    \mathcal{L}^{l}_{\mathrm{target}}
    =
    -
    \log (\bar{s}_{\mathcal{T}}^{l} + \varepsilon),
    \label{eq:decoy_strength_loss}
\end{equation}
where the logarithm increases the sensitivity to small attention values, which are typically tightly clustered due to the softmax normalization, while $\varepsilon$ is introduced for numerical stability.
Minimizing this term increases the attention received by the target region at layer $l$.

\begin{figure}[!t]
\centering
\includegraphics[width=\columnwidth]{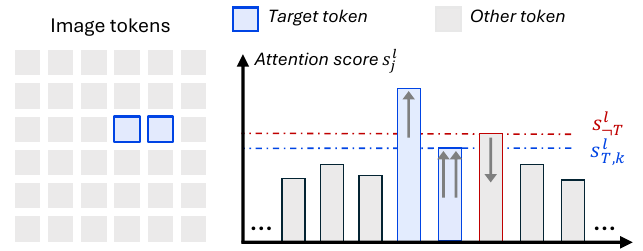}
\caption{\small{
Illustration of the decoy optimization objective: the attention of the target tokens (blue) is increased overall, with additional emphasis on the weakest target score of the top-$k$ ($k=2$), pushed to exceed the strongest non-target score (red).}}
\label{fig:loss_idea}
\end{figure}

Increasing the absolute attention received by the target region, however, does not necessarily ensure that its tokens appear among those selected by the defense. We therefore introduce a second objective that explicitly promotes their dominance in the received-attention ranking. Let
\(
    s_{\mathcal{T},k}^{l}
    =
    \min_{i\in
    \operatorname{Top_k}(\mathcal{T},l)}
    s_i^l
    \label{eq:topk_target_attention}
\)
denote the \(k\)-th largest attention score within \(\mathcal{T}\). Here,
\(\operatorname{Top_k}(\mathcal{T},l)\) returns the indices of the \(k\) highest-scoring tokens in the target region.
The strongest competing score outside the target region is defined as
\(
    s_{\neg\mathcal{T}}^{l}
    =
    \max_{j\in
    \mathcal{I}\setminus \mathcal{T}}
    s_j^l.
    \label{eq:strongest_nontarget}
\)
The layer-wise target-dominance ratio is then
\begin{equation}
    R^l
    =
    \frac{
        s_{\mathcal{T},k}^{l}
    }{
        s_{\neg \mathcal{T}}^{l}
        +\varepsilon
    },
    \label{eq:decoy_ratio}
\end{equation}
where a value \(R^l>1\) indicates that even the weakest token among the \(k\) strongest target tokens
receives more attention than the strongest token outside the target region.
Therefore, the \(k\) strongest target tokens occupy the top \(k\) positions
of the attention ranking at layer \(l\).
Following this, a layer-wise ratio objective is defined as
\begin{equation}
    \mathcal{L}^l_{\mathrm{ratio}}
    =
    -
    \log(R^l + \varepsilon).
    \label{eq:decoy_ratio_loss}
\end{equation}
Minimizing this term increases the relative attention assigned to the target tokens with respect to the strongest non-target competitor. An illustration is shown in Fig.~\ref{fig:loss_idea}.

Although maximizing $R^l$ promotes the dominance of the target tokens, different layers may reach the desired ratio at different stages of the optimization (see App.\ref{appx:optimization_details}). Applying the same weight to all layers may therefore allocate excessive optimization effort to layers in which the target tokens already dominate, while giving insufficient emphasis to layers that have not yet reached the desired ranking. To account for this imbalance, we introduce a specific layer weight. Given a desired dominance ratio $r>1$, we define
\begin{equation}
    \alpha^l
    =
    \begin{cases}
        \alpha_0,
        & R^l < r,\\[1mm]
        \alpha_1,
        & R^l \geq r,
    \end{cases}
    \qquad
    \alpha_0
    \geq
    \alpha_1
    \geq 0.
    \label{eq:layer_status_weight}
\end{equation}
Layers that have not yet reached the desired ratio therefore receive a stronger optimization signal. Conversely, layers that already satisfy the dominance condition receive a smaller contribution, allowing the optimization to focus primarily on the remaining layers while still retaining some pressure on already satisfied ones when \(\alpha_1>0\).
The complete attention-based objective is then
\begin{equation}
\mathcal{L}_{\text{decoy}}
=
\frac{1}{|{L}_{\mathcal{D}}|}
\sum_{l\in {L}_{\mathcal{D}}}
\alpha^l
\left(
\mathcal{L}_{\mathrm{target}}^{l}
+
\beta_{\mathrm{ratio}}
\mathcal{L}_{\mathrm{ratio}}^{l}
\right),
\label{eq:overall_equation_decoy}
\end{equation}
where $\beta_{\mathrm{ratio}}\geq 0$ controls the importance of the ranking-dominance objective relative to the maximization of the overall attention received by the target region. 

Let \(p_{\mathcal{D}}\) denote the learnable pixel values associated with the modifiable decoy region and let \(\mathcal{P}_{\mathcal{D}}\) denote their feasible domain. The optimized decoy is obtained by solving
\begin{equation}
    p_{\mathcal{D}}^{\star}
    =
    \arg\min_{p_{\mathcal{D}}\in\mathcal{P}_{\mathcal{D}}}
    \mathcal{L}_{\mathrm{decoy}}.
    \label{eq:decoy_optimization}
\end{equation}


\subsection{Bypassing defenses with decoy patches}
\label{sec:bypassing_defenses}

The proposed bypass paradigm highlights the underlying vulnerability of defenses based on attention weights. It combines localized adversarial attacks with optimized decoy patches, without relying on a specific attack formulation. The complete procedure for a single image is summarized in Algorithm~\ref{alg:decoy_bypassing} and described below.



\begin{algorithm}[!t]
\caption{Attack with Decoys}
\label{alg:decoy_bypassing}
\DontPrintSemicolon

\algblockgreen{Phase 1 -- Adversarial Attack Optimization}

\(x_{\mathcal{A}}
\leftarrow
\operatorname{Attack}(x,y,f,\mathcal{A})\)\;

\algblockblue{Phase 2 -- Adversarial Decoy Optimization}

Initialize \(p_{\mathcal{D}}^{(0)}
\in\mathcal{P}_{\mathcal{D}}\)\;

\For{\(t=0,\ldots,\text{num\_steps}\)}{
    Construct \(x_{\mathcal{A},\mathcal{D}}^{(t)}\) by adding
    \(p_{\mathcal{D}}^{(t)}\) to 
    \(x_{\mathcal{A}}\)\;

    Compute \(\mathcal{L}_{\mathrm{decoy}}\) from
    \(x_{\mathcal{A},\mathcal{D}}^{(t)}\) over  layers \({L}_{\mathcal{D}}\)\;

    Update the decoy content:
    \[
        p_{\mathcal{D}}^{(t+1)}
        \leftarrow
        \Pi_{\mathcal{P}_{\mathcal{D}}}
        \left(
            p_{\mathcal{D}}^{(t)}
            -
            \eta
            \nabla_{p_{\mathcal{D}}^{(t)}}
            \mathcal{L}_{\mathrm{decoy}}
        \right);
    \]
}
\Return \(x_{\mathcal{A},\mathcal{D}}^{(t)}\)\;
\end{algorithm}

Algorithm~\ref{alg:decoy_bypassing} is divided into two phases, where two disjoint sets of image tokens are manipulated. The first phase performs the adversarial optimization and focuses on \(\mathcal{A}\subseteq\mathcal{I}\), which denotes the tokens associated with the adversarial region. Here, a localized attack \(\operatorname{Attack(\cdot)}\), which can be instantiated with different localized attack formulations, is applied within \(\mathcal{A}\) to generate adversarial content that induces misclassification.
The location of \(\mathcal{A}\) can either be specified manually or determined according to the adopted attack (e.g., ~\cite{fu2022patch}).
Once the adversarial example \( x_{\mathcal{A}} \)\ has been generated, the content of the adversarial region is kept fixed during the subsequent steps.
The second phase performs the decoy optimization on the adversarial input and focuses on \(\mathcal{D}\subseteq\mathcal{I}\), the tokens associated with the decoy region. This region can be selected manually or randomly, ensuring that \( \mathcal{A}\cap\mathcal{D}=\emptyset \). First, the decoys are initialized within the decoy region with random values. 
Then, they are optimized for \textit{num\_steps} iterations.
At every step, the decoy tokens are applied similarly to Eq.~\ref{eq:token_replacement}, and are optimized using a gradient-based procedure \cite{madry2018towards} with a step size $\eta$, and with the objective introduced in Sec.~\ref{sec:decoy_optimization} (\(\mathcal{L}_{\mathrm{decoy}}\)) over a specified set of layers \({L}_{\mathcal{D}}\). 
In our implementation, the target region coincides with the decoy patches location, i.e., \(\mathcal{T}=\mathcal{D}\), where every patch is aligned with the tokenization of ViTs, i.e., it corresponds to a token of the model, as in the main works discussed in this paper. In Algorithm~\ref{alg:decoy_bypassing}, \(\Pi_{\mathcal{P}_{\mathcal{D}}}\) denotes the projection of \(p_{\mathcal{D}}\) onto its admissible set, which may, for example, enforce the valid pixel range \([0,1]\). The resulting final input contains both the original adversarial content and the optimized decoy region. 
Separating the adversarial and decoy optimization into two phases increases the flexibility of the approach, allowing the decoy tokens to be combined with different localized attacks, as demonstrated in Sec.~\ref{sec:exp}.

\section{Experiments}
\label{sec:exp}

\subsection{Experimental setup}
\label{sec:exp_setup}
\noindent \textbf{Models and dataset.}
For testing the effectiveness of our methodology, we considered different ViT architectures:
DeiT-B/16-224~\cite{touvron2021training}, ViT-B/16-224, and ViT-S/16-224~\cite{dosovitskiy2020image}. 
All models are initialized from publicly available pretrained weights in \texttt{timm}\footnote{\url{https://github.com/rwightman/timm}}. Experiments are conducted on the ImageNet validation set~\cite{deng2009imagenet}. Following PatchFool and ARMRO, and due to the high cost of optimizing both adversarial and decoy regions (see App.~\ref{appx:optimization_details}), 
we employ a subset of the dataset consisting of $1024$ samples, randomly selected with a seed kept fixed across all experiments to ensure fairness of the results.

\noindent \textbf{Decoy setting.}
We evaluate decoys under different token configurations. In the main experiments, decoy locations overlap with tokens randomly selected from those excluding the adversarial tokens. Additional experiments on the effect of decoy placement are reported in App.~\ref{app:position_analysis}, where we observe negligible differences.
Decoys are optimized for $2500$ iterations, which we found necessary for the loss to converge across all ViT layers (see App.~\ref{appx:optimization_details}). Unless otherwise stated, we set the optimization parameters to $\beta_\mathrm{ratio}=10$ and the dominance threshold to $r=2$, $\alpha_0 = 1$, $\alpha_1 = 0.05$, $\eta = 0.05$, and the set \({L}_{\mathcal{D}}\) includes all layers. Ablations on the loss terms are in Sec.~\ref{sec:ablation}.

\noindent \textbf{Attack and defense setting.}
We combine adversarial decoys with two localized attacks. First, we consider PatchFool~\cite{fu2022patch}, a strong token-level attack specifically designed for ViTs.
Second, we evaluate a standard adversarial patch attack~\cite{brown2017adversarial}, which perturbs a contiguous region of the image. For PatchFool, we follow the settings of the original implementation. For the adversarial patch attack, we use a learning rate $0.05$. Both attacks are optimized for $500$ iterations and with a perturbation magnitude $1.0$ to simulate a classic adversarial patch. 
For each attack, we also introduce a \textit{mask-aware variant}, where the original attack objective is left unchanged, while random masking is injected during optimization to improve attack persistence against defense-induced sparse suppression.
This accounts for the fragility of localized attacks under masking, which can also occur when the defense does not cover the attacked regions. We report details and additional analysis of the mask-aware attacks in App.~\ref{app:mask_optimization}.
Finally, we consider, in Sec.~\ref{sec:budget_analysis}, an attention-aware attack, which jointly optimizes the task objective while explicitly discouraging attention on the attacked tokens, i.e.,
\(
\mathcal{L}_{\mathrm{adapt}}
=
\mathcal{L}_{\mathrm{cls}}(f(x_{\mathcal{A}}), y)
-
\lambda\frac{1}{|{L}||\mathcal{A}|}
\sum_{l\in{L}}
\sum_{j\in\mathcal{A}}
s_j^l ,
\)
where $\lambda=0.01$ controls the strength of the attention-aware term. More details of the optimization strategy are in App.~\ref{appx:adaptive_attack}.

As the main defense, we use ARMRO~\cite{liu2023understanding} due to its strong performance, as discussed in Sec.~\ref{sec:RelatedWorks}.
We mask pixels with the image mean as in the original paper and use ARMRO threshold $\tau=1.2$ for DeiT-B and ViT-B and $\tau=1.4$ for ViT-S, yielding high robust accuracy with a negligible drop in clean defended accuracy.
Additional evaluations against RSA and a Top-$K$ attention suppression defense are in App.~\ref{appx:additional_defense}.

\begin{table*}[t]
\centering
\scriptsize
\setlength{\tabcolsep}{2.5pt}


\newcommand{\acc}[2]{%
    \ensuremath{#1{\scriptscriptstyle\,\pm\,#2}}%
}

\newcommand{\deltaacc}[1]{%
    \textcolor{blue!65!black}{%
        \bfseries\ensuremath{#1}%
    }%
}

\newcommand{\rowstrut}{\rule{0pt}{2.5ex}}

\newcommand{\ResultRow}[7]{%
    #1
    & \rowstrut #2
    & #3
    & #4
    & #5
    & #6 \\
}

\begin{tabular}{@{}c@{\hspace{0.02\textwidth}}c@{}}

\begin{tabular}{c}
\label{tab:main_results_a}
\textbf{(a) Standard attack} \\[2pt]

\resizebox{0.48\textwidth}{!}{%
\begin{tabular}{@{}c|c|c|c|c|c|c@{}}

\toprule
\multicolumn{7}{c}{PatchFool~\cite{fu2022patch}} \\
\midrule

\shortstack{Model} &
\shortstack{\#Adv.T} &
\shortstack{Clean} &
\shortstack{Att} &
\shortstack{Att+Dec} &
\shortstack{Att+Def} &
\shortstack{Att+Dec+Def} \\

\midrule


\multirow{3}{*}{\textbf{DeiT}}
& 1
& 82.32
& \acc{14.45}{1.10}
& \acc{68.26}{1.46}
& \acc{{80.96}}{1.23}
& \acc{52.34}{1.56} \\

& 2
& -
& \acc{0.68}{0.26}
& \acc{61.13}{1.52}
& \acc{80.08}{1.25}
& \acc{42.09}{1.54} \\

& 4
& -
& \acc{0.00}{0.00}
& \acc{55.18}{1.55}
& \acc{77.05}{1.31}
& \acc{26.86}{1.39} \\

\midrule


\multirow{3}{*}{\textbf{ViT-B}}
& 1
& 85.06
& \acc{26.76}{1.38}
& \acc{66.89}{1.47}
& \acc{84.18}{1.14}
& \acc{{53.03}}{1.56} \\

& 2
& -
& \acc{4.10}{0.62}
& \acc{58.01}{1.54}
& \acc{81.35}{1.22}
& \acc{31.05}{1.45} \\

& 4
& -
& \acc{0.00}{0.00}
& \acc{40.33}{1.53}
& \acc{76.07}{1.33}
& \acc{{20.12}}{1.25} \\

\midrule


\multirow{3}{*}{\textbf{ViT-S}}
& 1
& 81.64
& \acc{10.94}{0.98}
& \acc{51.46}{1.56}
& \acc{77.93}{1.30}
& \acc{{27.64}}{1.40}\\

& 2
& -
& \acc{0.98}{0.31}
& \acc{38.87}{1.52}
& \acc{71.48}{1.41}
& \acc{18.46}{1.21} \\

& 4
& -
& \acc{0.20}{0.14}
& \acc{22.36}{1.30}
& \acc{58.79}{1.54}
& \acc{10.55}{0.96} \\


\midrule
\multicolumn{7}{c}{%
    Adversarial Patch~\cite{brown2017adversarial}
} \\
\midrule


\multirow{2}{*}{\textbf{DeiT}}
& 1
& 82.32
& \acc{18.16}{1.21}
& \acc{59.57}{1.53}
& \acc{79.79}{1.26}
& \acc{50.49}{1.56} \\

& 4
& -
& \acc{4.49}{0.65}
& \acc{45.70}{1.56}
& \acc{73.05}{1.39}
& \acc{29.30}{1.42} \\

\midrule


\multirow{2}{*}{\textbf{ViT-B}}
& 1
& 85.06
& \acc{18.95}{1.23}
& \acc{59.08}{1.54}
& \acc{83.50}{1.16}
& \acc{{42.58}}{1.55} \\

& 4
& -
& \acc{0.49}{0.22}
& \acc{40.04}{1.53}
& \acc{78.52}{1.28}
& \acc{17.48}{1.19} \\

\midrule


\multirow{2}{*}{{\textbf{ViT-S}}}
& 1
& 81.64
& \acc{8.30}{0.86}
& \acc{48.44}{1.56}
& \acc{76.07}{1.33}
& \acc{25.29}{1.36} \\

& 4
& -
& \acc{0.00}{0.00}
& \acc{24.90}{1.35}
& \acc{{63.77}}{1.50}
& \acc{11.72}{1.01} \\

\bottomrule
\end{tabular}%
}

\end{tabular}

\begin{tabular}{c}
\label{tab:main_results_b}
\textbf{(b) Mask-aware optimization} \\[2pt]

\resizebox{0.48\textwidth}{!}{%
\begin{tabular}{@{}c|c|c|c|c|c|c@{}}

\toprule
\multicolumn{7}{c}{PatchFool~\cite{fu2022patch}} \\
\midrule

\shortstack{Model} &
\shortstack{\# Adv.T} &
\shortstack{Clean} &
\shortstack{Att} &
\shortstack{Att+Dec} &
\shortstack{Att+Def} &
\shortstack{Att+Dec+Def} \\

\midrule


\multirow{3}{*}{\textbf{DeiT}}
& 1
& 82.32
& \acc{33.79}{1.48}
& \acc{62.70}{1.51}
& \acc{80.08}{1.25}
& \acc{{50.29}}{1.56} \\

& 2
& -
& \acc{4.00}{0.61}
& \acc{50.68}{1.56}
& \acc{{78.12}}{1.29}
& \acc{{28.32}}{1.41} \\

& 4
& -
& \acc{0.00}{0.00}
& \acc{37.01}{1.51}
& \acc{{68.26}}{1.46}
& \acc{{11.13}}{0.98} \\

\midrule


\multirow{3}{*}{\textbf{ViT-B}}
& 1
& 85.06
& \acc{35.74}{1.50}
& \acc{64.55}{1.50}
& \acc{{83.50}}{1.16}
& \acc{52.25}{1.56} \\

& 2
& -
& \acc{8.30}{0.86}
& \acc{53.52}{1.56}
& \acc{{80.47}}{1.24}
& \acc{{29.30}}{1.42} \\

& 4
& -
& \acc{0.10}{0.10}
& \acc{34.38}{1.48}
& \acc{{66.89}}{1.47}
& \acc{14.65}{1.11} \\

\midrule


\multirow{3}{*}{\textbf{ViT-S}}
& 1
& 81.64
& \acc{14.94}{1.11}
& \acc{48.54}{1.56}
& \acc{76.66}{1.32}
& \acc{30.96}{1.45} \\

& 2
& -
& \acc{1.46}{0.38}
& \acc{35.45}{1.50}
& \acc{{67.77}}{1.46}
& \acc{17.68}{1.19} \\

& 4
& -
& \acc{0.00}{0.00}
& \acc{13.57}{1.07}
& \acc{{41.80}}{1.54}
& \acc{{6.15}}{0.75} \\


\midrule
\multicolumn{7}{c}{%
    Adversarial Patch~\cite{brown2017adversarial}
} \\
\midrule


\multirow{2}{*}{\textbf{DeiT}}
& 1
& 82.32
& \acc{28.52}{1.41}
& \acc{57.62}{1.55}
& \acc{79.49}{1.26}
& \acc{41.99}{1.54} \\

& 4
& -
& \acc{1.27}{0.35}
& \acc{47.07}{1.56}
& \acc{{74.22}}{1.37}
& \acc{{21.58}}{1.29}\\

\midrule


\multirow{2}{*}{\textbf{ViT-B}}
& 1
& 85.06
& \acc{23.93}{1.33}
& \acc{54.79}{1.56}
& \acc{83.01}{1.17}
& \acc{46.88}{1.56}\\

& 4
& -
& \acc{0.10}{0.10}
& \acc{30.76}{1.44}
& \acc{{72.75}}{1.39}
& \acc{17.58}{1.19} \\

\midrule


\multirow{2}{*}{\textbf{ViT-S}}
& 1
& 81.64
& \acc{8.40}{0.87}
& \acc{48.93}{1.56}
& \acc{{77.05}}{1.31}
& \acc{{24.90}}{1.35} \\

& 4
& -
& \acc{0.00}{0.00}
& \acc{23.54}{1.33}
& \acc{65.33}{1.49}
& \acc{10.94}{0.98} \\

\bottomrule
\end{tabular}%
}
\end{tabular}

\end{tabular}

\caption{\small
Accuracy under PatchFool and Adversarial Patch attacks, with $1$, $2$, $4$ and with $1$, $4$ adversarial tokens respectively, for the attack-only setting (\emph{Att}), with $4$ adversarial decoys (\emph{Att+Dec}), under ARMRO defense (\emph{Att+Def}), and with both decoys and ARMRO (\emph{Att+Dec+Def}). Subtables (a) and (b) report evaluations using standard and mask-aware attack optimization, respectively. Accuracy is reported as the mean, with $\pm$ indicating the standard error across the evaluated data.
\emph{Clean} denotes the mean clean accuracy of the model, while \emph{\#Adv.T} denotes the number of adversarial tokens used by the attack.
}
\label{tab:main_results}
\end{table*}

\begin{figure}[t]
    \centering
    \begin{subfigure}[t]{0.49\textwidth}
    \begin{subfigure}[t]{\columnwidth}
    \begin{subfigure}[t]{0.24\columnwidth}
        \centering
        \caption*{\footnotesize{Attack}}
    \includegraphics[width=\linewidth]{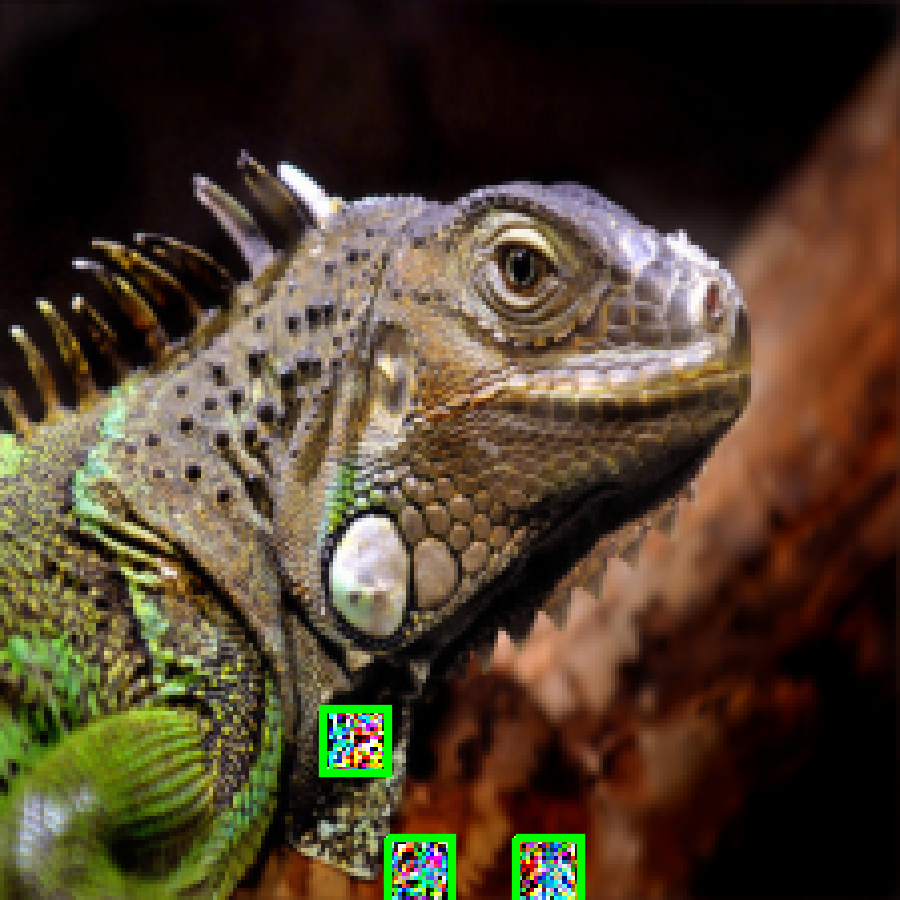}
    \end{subfigure}
    \hfill
    \begin{subfigure}[t]{0.24\columnwidth}
        \centering
        \caption*{\footnotesize{Att.$@$Layer}5}
    \includegraphics[width=\linewidth]{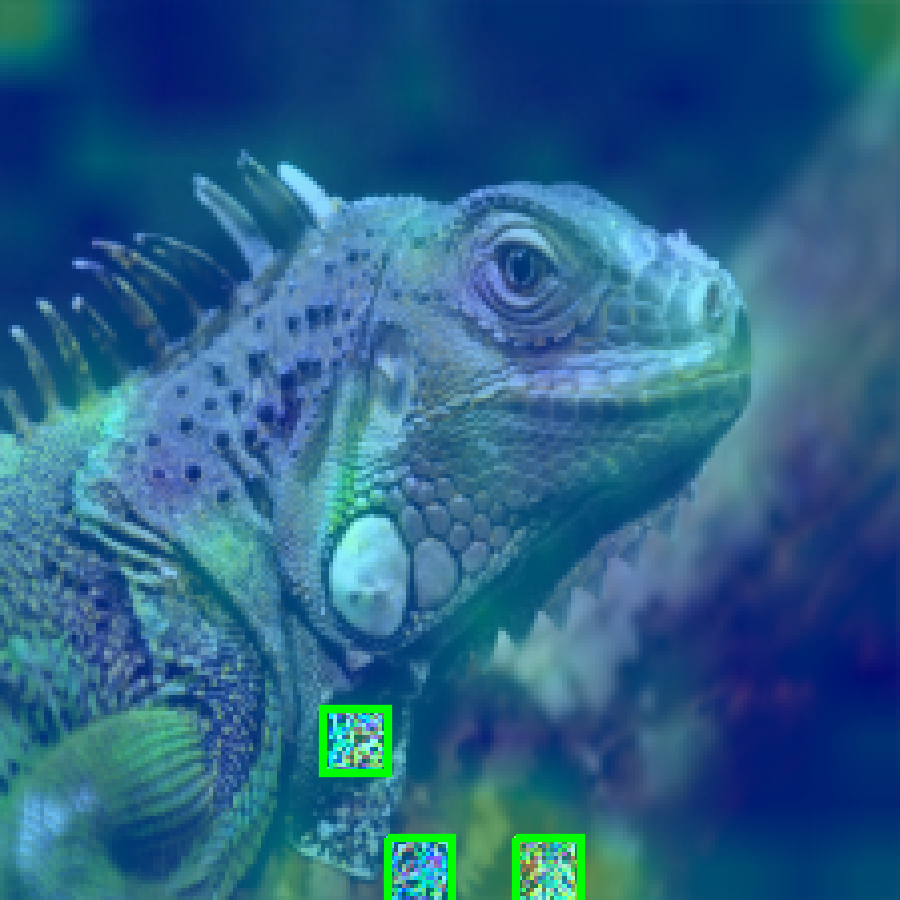}
    \end{subfigure}
    \hfill
    \begin{subfigure}[t]{0.24\columnwidth}
        \caption*{\footnotesize{Att.$@$Layer12}}
        \centering
    \includegraphics[width=\linewidth]{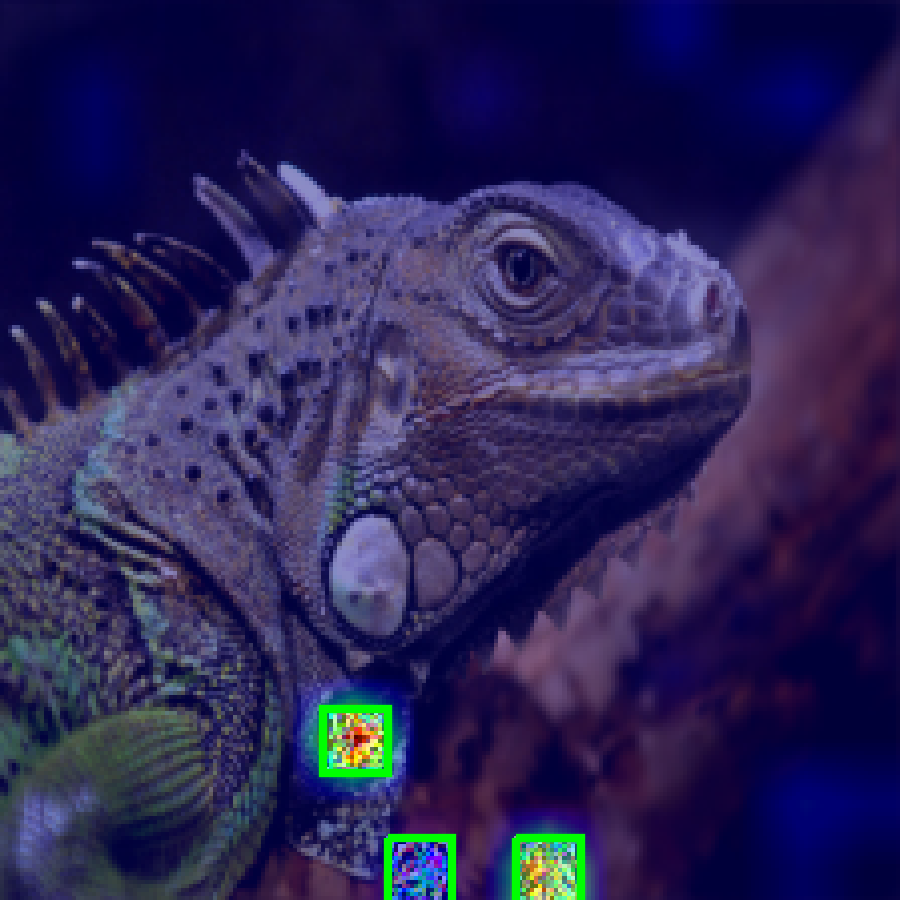}
    \end{subfigure}
    \hfill
    \begin{subfigure}[t]{0.24\columnwidth}
        \centering
        \caption*{\footnotesize{w/ defense}}
    \includegraphics[width=\linewidth]{{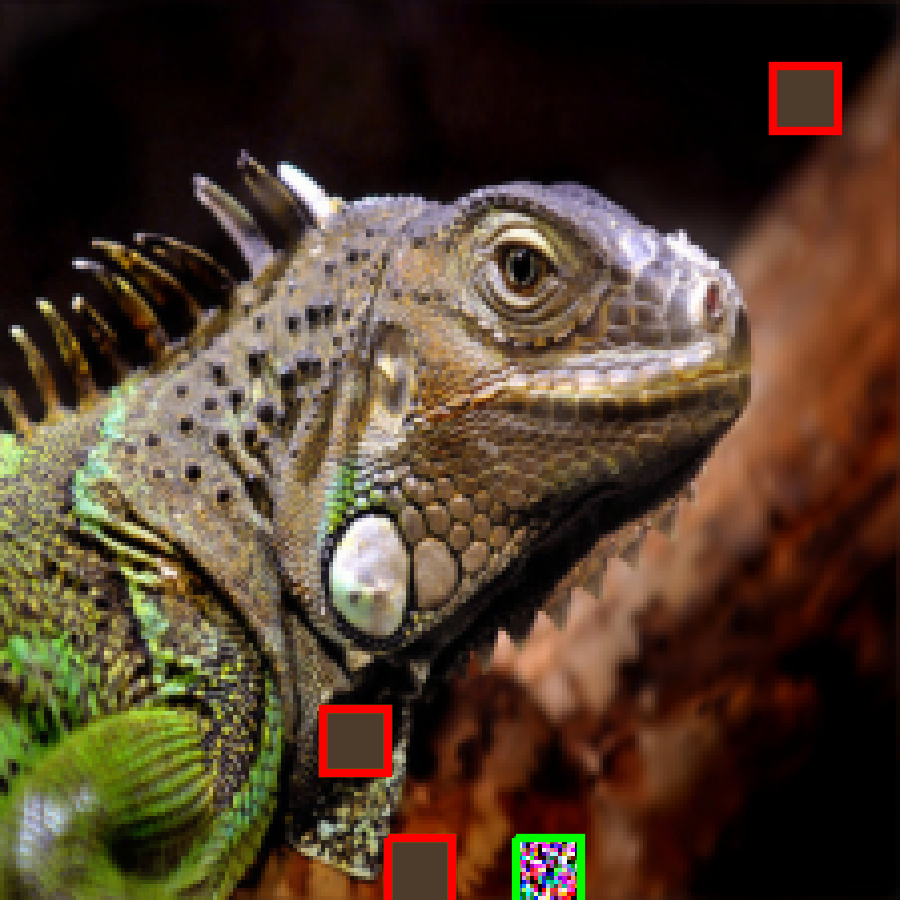}}
    \end{subfigure}
    \end{subfigure}
    \begin{subfigure}[t]{\columnwidth}
    \begin{subfigure}[t]{0.24\columnwidth}
        \centering
    \includegraphics[width=\linewidth]{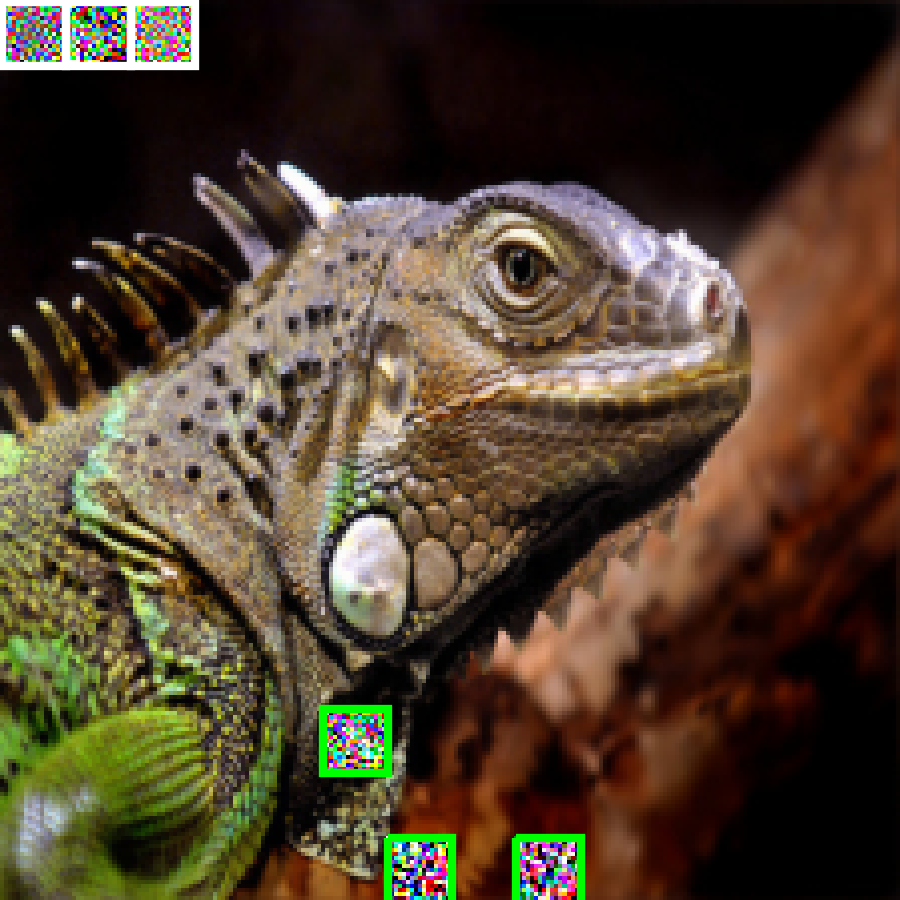}
    \end{subfigure}
    \hfill
    \begin{subfigure}[t]{0.24\columnwidth}
        \centering
    \includegraphics[width=\linewidth]{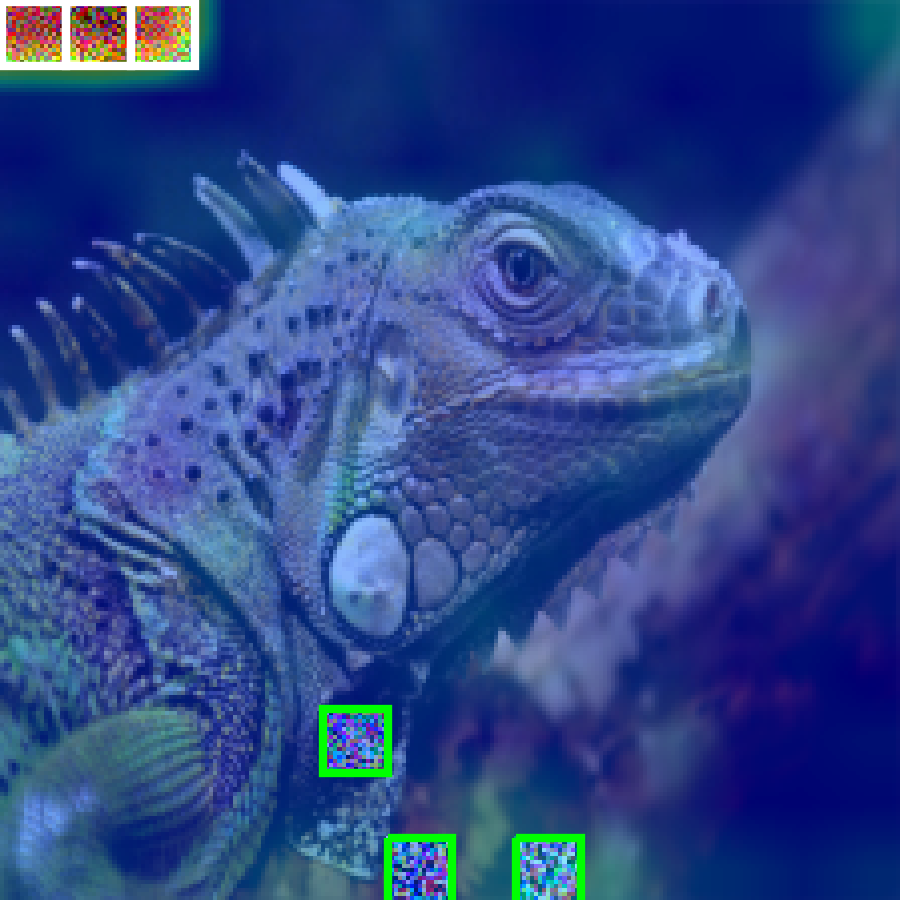}
    \end{subfigure}
    \hfill
    \begin{subfigure}[t]{0.24\columnwidth}
        \centering
    \includegraphics[width=\linewidth]{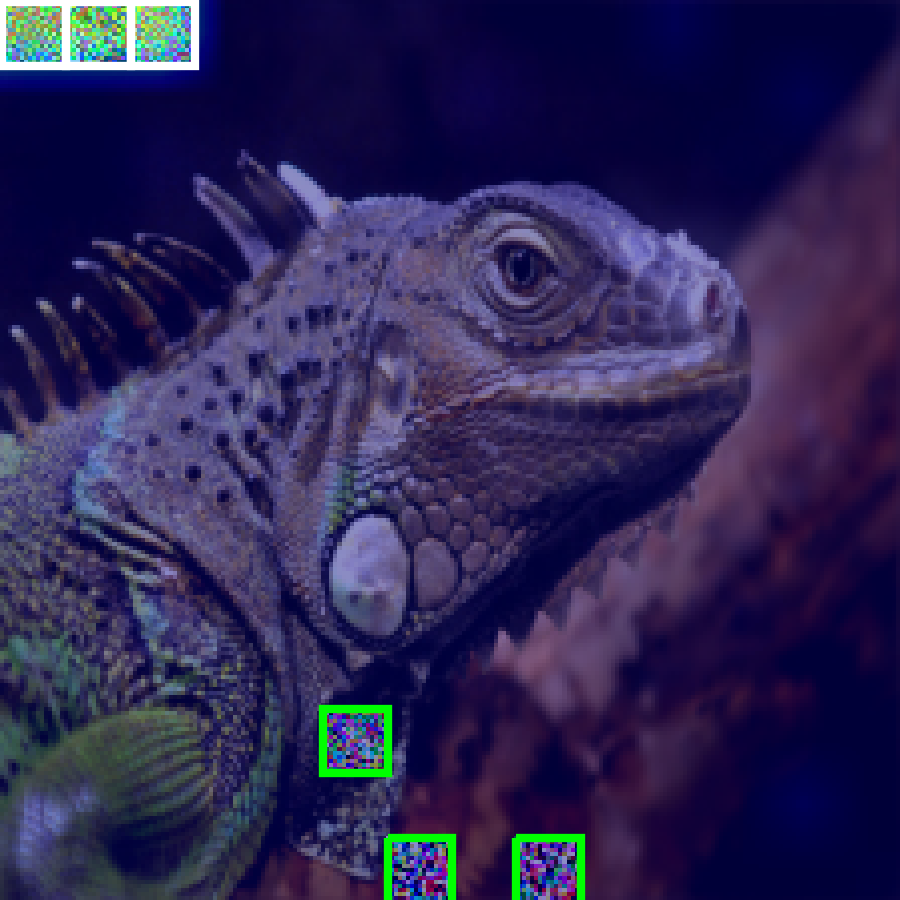}
    \end{subfigure}
    \hfill
    \begin{subfigure}[t]{0.24\columnwidth}
        \centering
    \includegraphics[width=\linewidth]{{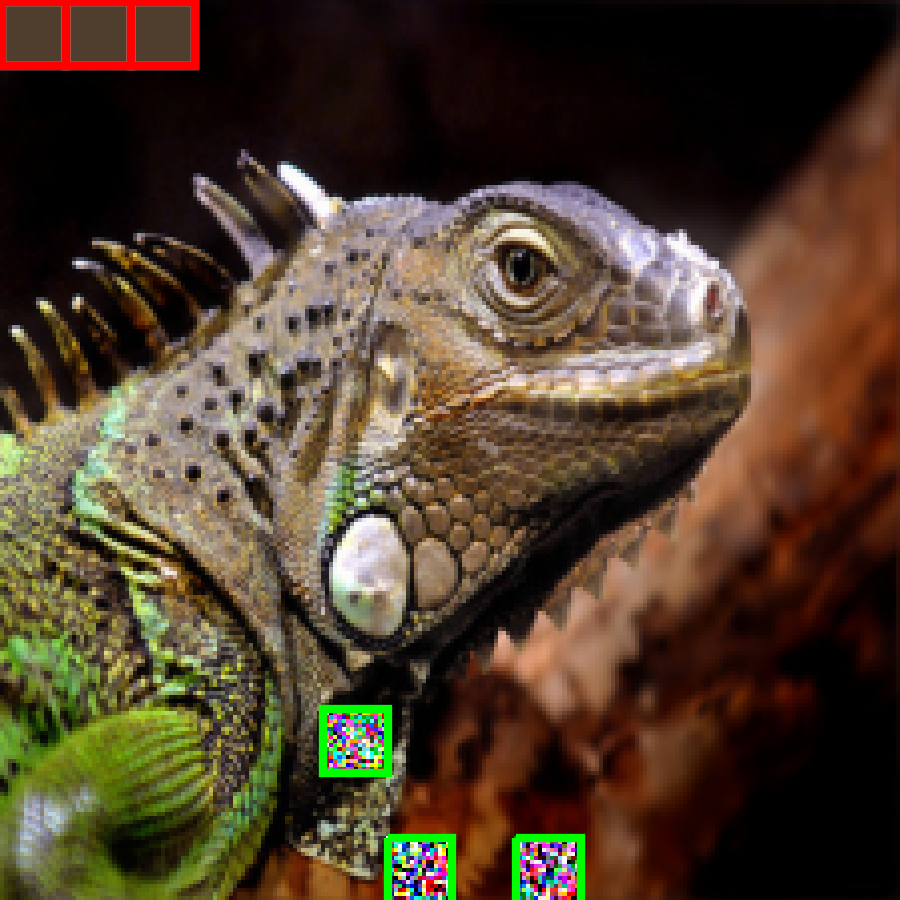}}
    \end{subfigure}
    \end{subfigure}
    \end{subfigure}
    \caption{\small{ImageNet qualitative example. Adversarial patches only on top, adversarial and decoy patches on the bottom. Adversarial patches in green, decoys in white, and covered patches in red.}}
    \label{fig:qualitative_results}
\end{figure}
\subsection{Main results}
\label{sec:main_results}
\rev{
Tab.~\ref{tab:main_results} reports the main results for PatchFool and adversarial patch attacks combined with adversarial decoys and ARMRO. We consider commonly used attack-token budgets, using $1$, $2$, and $4$ tokens for sparse PatchFool and $1$ and $4$ for the adversarial patch, implemented as a single squared patch, balancing attack visibility and effectiveness. 
Both the standard attack optimization setting (left) and the mask-aware variant (right) have been evaluated.
The number of decoy tokens here was set to $4$, since it provided the best results within a limited spatial budget. Experiments with different patch budgets are investigated next. 
}

The results show that ARMRO is effective against localized attacks in the absence of decoys (\textit{Att+Def}), often restoring accuracy close to the clean level. However, this recovery is substantially reduced when decoys are introduced (\textit{Att+Dec+Def}), with defended accuracy dropping significantly across architectures, attacks, and attack sizes. In the absence of the defense (\textit{Att+Dec}), decoys reduce the attack effectiveness, as expected since they are designed to interact with defense-induced masking rather than reinforce the attack itself. 
These behaviors are consistent across both attack types, supporting the claim that the bypassing pipeline generalizes across different attack formulations.
Note that decoys are optimized to shift attention, without any adversarial objective (further details are provided in App.~\ref{appx:adverarial_effect_decoy}).

Fig.~\ref{fig:qualitative_results} provides a qualitative illustration of the decoy mechanism for defense misdirection using $3$ decoy tokens combined with $3$ PatchFool tokens.

\subsection{Number of decoys and masking coverage}
\label{sec:num_decoys}
\begin{figure*}[t]
    \centering

    \hfill
    \begin{subfigure}{\textwidth}
        \centering
        \includegraphics[width=0.40\linewidth]{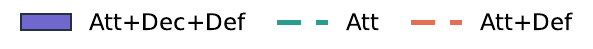}
    \end{subfigure}
    \hfill
    \begin{subfigure}[t]{0.24\textwidth}
        \centering
        \includegraphics[width=\linewidth]{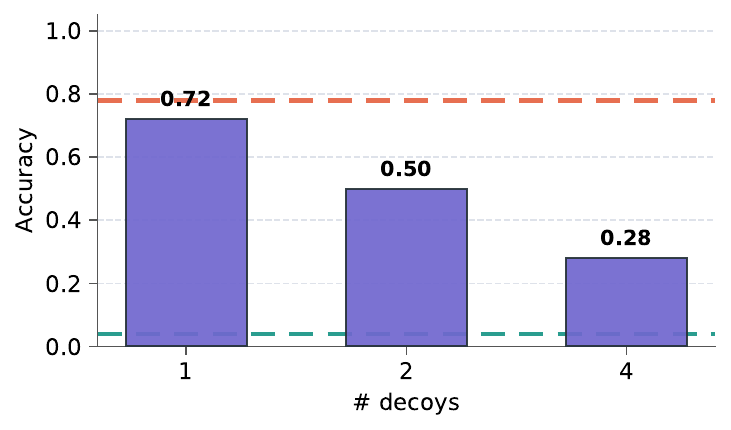}
    \end{subfigure}
    \hfill
    \begin{subfigure}[t]{0.24\textwidth}
        \centering
        \includegraphics[width=\linewidth]{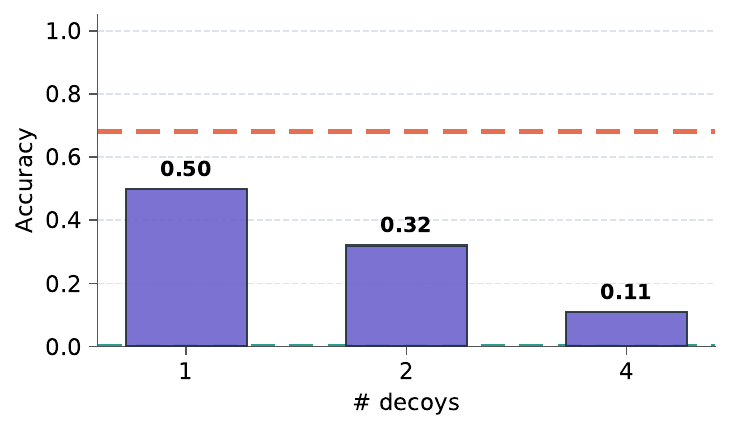}
    \end{subfigure}
    \hfill
    \begin{subfigure}[t]{0.24\textwidth}
        \centering
        \includegraphics[width=\linewidth]{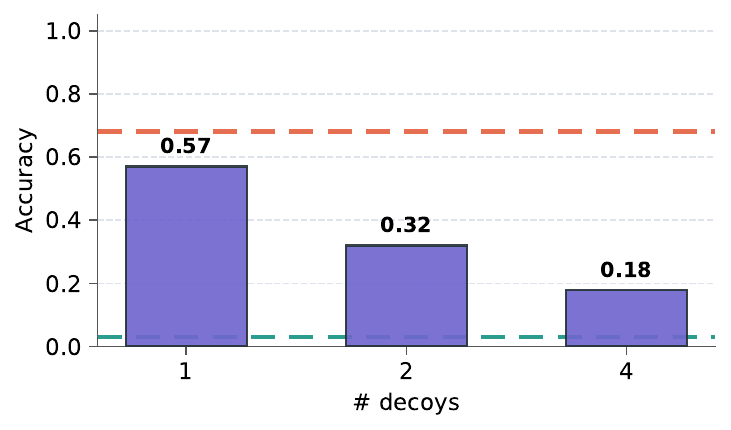}
    \end{subfigure}
    \hfill
    \begin{subfigure}[t]{0.24\textwidth}
        \centering
        \includegraphics[width=\linewidth]{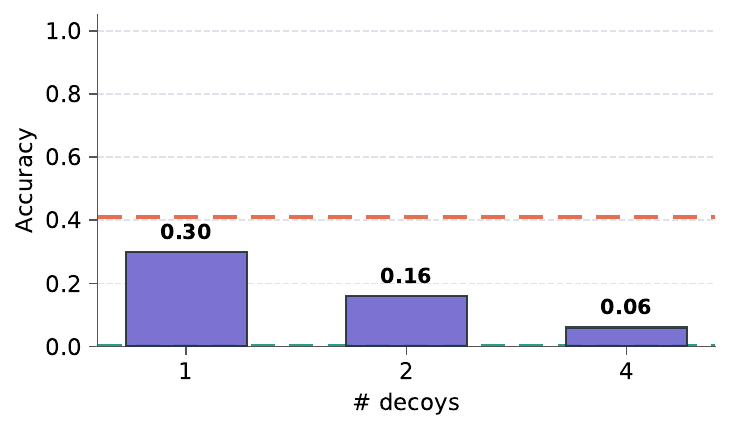}
    \end{subfigure}
    \hfill
    \begin{subfigure}{\textwidth}
        \centering
        \includegraphics[width=0.6\linewidth]{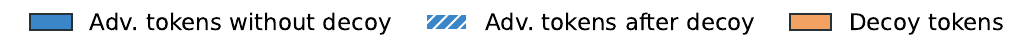}
    \end{subfigure}
    \hfill
    \begin{subfigure}[t]{0.24\textwidth}
        \centering
        \includegraphics[width=\linewidth]{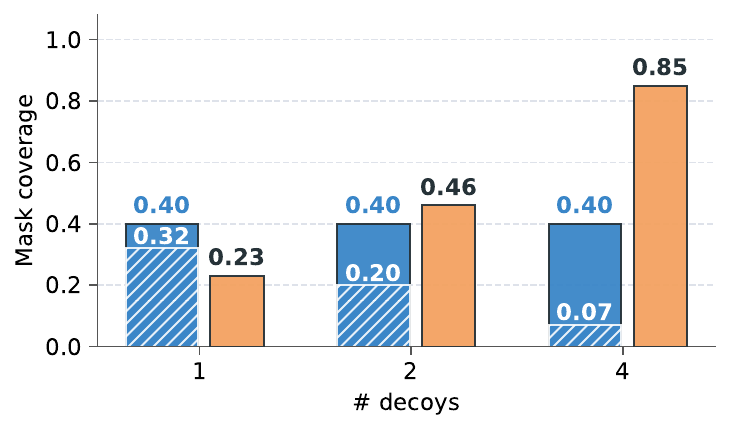}
        \caption{DeiT - \#PF-mask Tokens 2}
        \label{fig:acc_cov_deit_2}
    \end{subfigure}
    \hfill
    \begin{subfigure}[t]{0.24\textwidth}
        \centering
        \includegraphics[width=\linewidth]{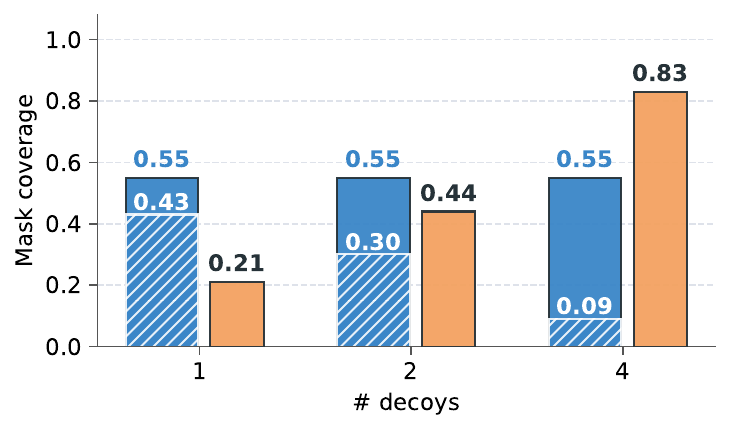}
        \caption{DeiT - \#PF-mask Tokens 4}
        \label{fig:acc_cov_deit_4}
    \end{subfigure}
    \hfill
    \begin{subfigure}[t]{0.24\textwidth}
        \centering
        \includegraphics[width=\linewidth]{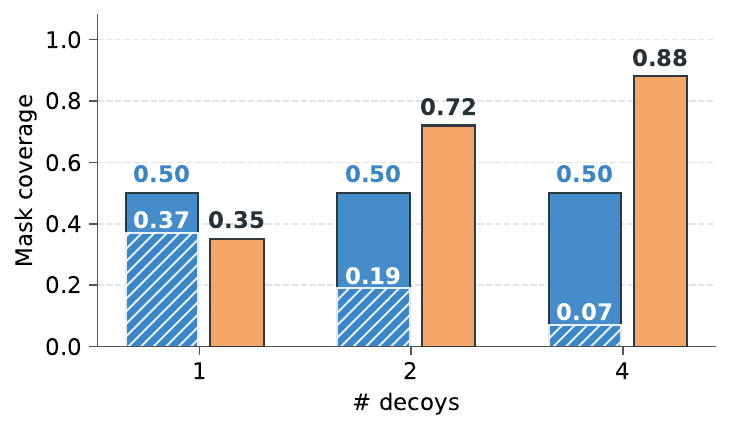}
        \caption{ViT-S - \#PF-mask Tokens 2}
        \label{fig:acc_cov_vits_2}
    \end{subfigure}
    \hfill
    \begin{subfigure}[t]{0.24\textwidth}
        \centering
        \includegraphics[width=\linewidth]{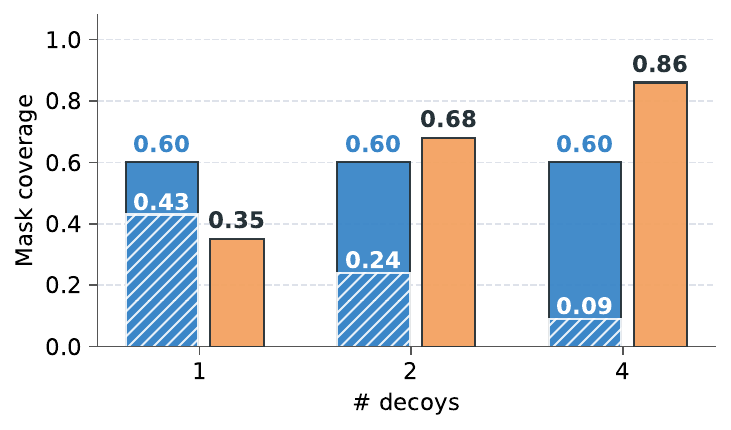}
        \caption{ViT-S - \#PF-mask Tokens 4}
        \label{fig:acc_cov_vits_4}
    \end{subfigure}
    \caption{\rev{\small{Effect of the number of decoys ($1$, $2$, $4$) on defense bypassing, for DeiT (a, b) and ViT-S (c, d) with attacked tokens fixed to $2$ (a, c) and $4$ (b, d). \emph{Top}: accuracy of \emph{Att+Dec+Def} (bars) vs.\ the undefended attack \emph{Att} (green dashed line) and the defended attack without decoys \emph{Att+Def} (red dashed line). \emph{Bottom}: defense coverage of adversarial tokens under \emph{Att+Def} (blue), compared to coverage of adversarial tokens (hatched blue) and decoy tokens (orange) under \emph{Att+Dec+Def}.}}}
\label{fig:complete_results}
\end{figure*}

Fig.~\ref{fig:complete_results} shows the effect of varying the number of decoy tokens for DeiT \rev{(\ref{fig:acc_cov_deit_2}, \ref{fig:acc_cov_deit_4})} and ViT-S \rev{(\ref{fig:acc_cov_vits_2}, \ref{fig:acc_cov_vits_4})}, while keeping the number of adversarial tokens fixed to $2$ \rev{(\ref{fig:acc_cov_deit_2}, \ref{fig:acc_cov_vits_2})} and $4$ patches \rev{(\ref{fig:acc_cov_deit_4}, \ref{fig:acc_cov_vits_4})}.
\rev{The first row (top) reports the robust accuracy after ARMRO for different decoy budgets. The dashed lines provide reference values for the attack alone (green) and for the attack combined with ARMRO, but without decoys (red). The second row (bottom) shows the defense mask coverage of the adversarial and decoy tokens under the same settings, i.e., the ratio of tokens of each type that are masked by the defense.}
Increasing the number of decoys generally improves the bypass: the defended accuracy decreases as more decoy tokens are introduced. The coverage analysis \rev{confirms} this behavior: with more decoys, the defense increasingly masks decoy tokens, while the coverage of the true adversarial tokens after adding the decoys drops substantially. Thus, decoys act by competing with the adversarial region for the attention-based suppression mask.

This also reveals a trade-off, where \rev{exploiting more} decoys improves defense bypassing, but increases the total number of manipulated tokens and may make the perturbation more visible. \rev{This trade-off is studied next.}

\begin{figure}[t]
    \centering
    \begin{subfigure}[t]{0.23\textwidth}
    \includegraphics[width=\linewidth]{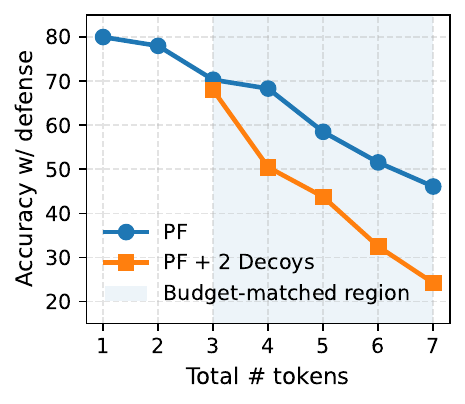}
    \end{subfigure}
    \begin{subfigure}[t]{0.23\textwidth}
    \includegraphics[width=\linewidth]{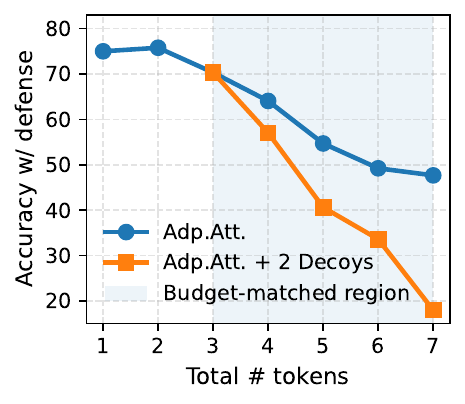}
    \end{subfigure}
    \vspace{-1em}
    \caption{\small{Patch-budget comparison on DeiT. Full budget on attack depicted in blue, $2$ patches reserved for decoys depicted in orange. PF-mask on the left, and Adaptive attack on the right.}}
    \label{fig:budget_analysis}
\end{figure}
\subsection{Patch-budget analysis and adaptive attack}
\label{sec:budget_analysis}
Although decoys are effective, they introduce additional tokens to manipulate.
Therefore, in this subsection, we evaluate whether reserving tokens for decoys is more effective than just allocating all the spatial resources to the adversarial attack.
Results are in Fig.~\ref{fig:budget_analysis}, where we compare two strategies: the attacker uses all available tokens as adversarial ones (blue line), and a decoy-based strategy that reserves two tokens for decoys (orange line). On the left, PatchFool is shown, while on the right, the attention-aware attack introduced in Sec.~\ref{sec:exp_setup}. The shaded area denotes where part of the token budget is allocated to the decoy.

In both settings, allocating part of the patches to decoys consistently yields lower defended accuracy, and the gap increases with the total budget. Thus, using decoys can be more beneficial than focusing only on the adversarial region. On average, in the shaded region, the attention-aware attack achieves $57.19\%$ accuracy without decoys and $43.90\%$ with decoys, while mask-aware PatchFool achieves $58.95\%$ and $43.78\%$, respectively. Comparisons for other models are provided in App.~\ref{appx:adaptive_attack}.

Beyond the patch budget, in App.~\ref{appx:optimization_details}, we analyze the optimization cost of the decoys and the iterations required to achieve the desired objective. While decoys provide an effective paradigm for evading defenses, we acknowledge that they require more optimization iterations than adversarial tokens. Consequently, comparing the two approaches under an identical iteration budget would either under-optimize the decoys or waste iterations on the attack.



\subsection{Ablation study}
\label{sec:ablation}
The previous experiments evaluate the complete attack-defense pipeline. Here, we instead isolate the decoy optimization and study the contribution of its loss components. Since the objective is to make decoy tokens dominate the attention ranking across layers, we measure how often they appear among the top-$k$ attention scores of each layer, using $k=4$, equal to the number of decoys.

Tab.~\ref{tab:threshold_ablation} evaluates the ratio-based layer regulation term (Eq.~\ref{eq:layer_status_weight}), which down-weights layers that have reached the threshold $r$. We report the top-$k$ decoy coverage on each layer, with the row mean shown in \emph{Avg}. This value is consistently higher with $r$-Reg, indicating that it prevents optimization from concentrating on only a subset of layers.

Tab.~\ref{tab:ratio_ablation} shows the influence of the ratio dominance term (Eq.\ref{eq:overall_equation_decoy}). Without it, decoys may still achieve high attention in some layers but rarely dominate the full top-$k$ ranking, whereas in the other case, full top-$k$ coverage is achieved in more layers.
Despite these positive results, forcing decoy tokens to reach the highest attention scores remains challenging in some layers, especially for ViT-B. We discuss this behavior in more detail in App.~\ref{appx:top_k_analysis}.

For completeness, App.~\ref{appx:adverarial_effect_decoy} investigates the absence of adversarial effects from decoys in isolation, supporting the analysis in Sec.~\ref{sec:num_decoys}. Furthermore, App.~\ref{app:position_analysis} studies the effect of different decoy placements.

\begin{table}[!t]
\centering
\small
\setlength{\tabcolsep}{3.2pt}
\renewcommand{\arraystretch}{1.05}
\resizebox{\columnwidth}{!}{%
\begin{tabular}{l|c|cccccccccccc|c}
\hline
Model & $r$-Reg. & L1 & L2 & L3 & L4 & L5 & L6 & L7 & L8 & L9 & L10 & L11 & L12 & Avg. \\
\hline
\multirow{2}{*}{DeiT-B}
& \cmark & \textbf{89} & 42 & 74 & \textbf{100} & \textbf{100} & \textbf{100} & \textbf{98} & \textbf{96} & \textbf{86} & \textbf{91} & \textbf{96} & \textbf{100} & \textbf{89.33} \\
& \xmark & 24 & \textbf{81} & \textbf{84} & \textbf{100} & 98 & 91 & 59 & 56 & 19 & 28 & 45 & 99 & 65.33 \\
\hline
\multirow{2}{*}{ViT-B}
& \cmark & \textbf{29} & \textbf{85} & \textbf{89} & \textbf{100} & \textbf{96} & \textbf{85} & \textbf{49} & \textbf{52} & \textbf{17} & \textbf{18} & \textbf{35} & \textbf{99} & \textbf{62.83} \\
& \xmark & 21 & 79 & 84 & \textbf{100} & \textbf{96} & 83 & 45 & 46 & \textbf{17} & \textbf{18} & 31 & \textbf{99} & 59.92 \\
\hline
\multirow{2}{*}{ViT-S}
& \cmark & \textbf{88} & \textbf{97} & \textbf{93} & \textbf{100} & \textbf{100} & \textbf{100} & \textbf{96} & \textbf{43} & \textbf{42} & \textbf{46} & \textbf{64} & 99 & \textbf{80.67} \\
& \xmark & 76 & 95 & 87 & \textbf{100} & \textbf{100} & \textbf{100} & 93 & 30 & 34 & 36 & 63 & \textbf{100} & 76.17 \\
\hline
\end{tabular}
}
\caption{\small{Ablation of the layer-wise regulation term ($r$-Reg.), showing top-$k$ decoy percentage coverage per layer. Best values in bold.}}
\label{tab:threshold_ablation}
\end{table}

\begin{table}[!t]
\centering
\scriptsize
\setlength{\tabcolsep}{3.0pt}
\renewcommand{\arraystretch}{1.05}
\resizebox{\columnwidth}{!}{%
\begin{tabular}{l|c|cccccccccccc|cc}
\hline
Model & Ratio & L1 & L2 & L3 & L4 & L5 & L6 & L7 & L8 & L9 & L10 & L11 & L12 & \#95 & \#100 \\
\hline
\multirow{2}{*}{DeiT-B}
& \cmark & 89 & 42 & 74 & \textbf{100} & \textbf{100} & \textbf{100} & 98 & 96 & 86 & 91 & 96 & \textbf{100} & \textbf{7} & \textbf{4} \\
& \xmark & 39 & 29 & 42 & 87 & 95 & 96 & 86 & 81 & 74 & 71 & 66 & 84 & 2 & 0 \\
\hline
\multirow{2}{*}{ViT-B}
& \cmark & 29 & 85 & 89 & \textbf{100} & 96 & 85 & 49 & 52 & 17 & 18 & 35 & 99 & \textbf{3} & \textbf{1} \\
& \xmark & 41 & 38 & 65 & 93 & 90 & 81 & 67 & 68 & 55 & 55 & 57 & 81 & 0 & 0 \\
\hline
\multirow{2}{*}{ViT-S}
& \cmark & 88 & 97 & 93 & \textbf{100} & \textbf{100} & \textbf{100} & 96 & 43 & 42 & 46 & 64 & 99 & \textbf{6} & \textbf{3} \\
& \xmark & 25 & 36 & 56 & 77 & 90 & 92 & 87 & 85 & 83 & 81 & 81 & 84 & 0 & 0 \\
\hline
\end{tabular}%
}
\caption{\small{Ablation of the dominance ratio term, showing top-$k$ decoy coverage per layer. We summarize the number of layers with scores $\ge95$ and $100$ in the last columns.}}
\label{tab:ratio_ablation}
\end{table}



\section{Conclusions}
This work introduced \emph{adversarial decoys}, patches optimized to manipulate the attention distribution. Decoys can be combined with adversarial attacks to decouple attention manipulation from the adversarial objective, misdirecting attention-based defenses while preserving much of the attack effectiveness. Analysis and results show that attention magnitude and adversarial effectiveness are not always aligned, and decoys can weaken attention-weight-based masking defenses across different adversarial patch attacks, tested against the state-of-the-art defense ARMRO~\cite{liu2023understanding}.
The findings of this paper highlight limitations of adversarial defenses in ViTs and motivate future work on designing new defenses robust to attention distribution shifts, as well as applying decoys to explainability methods and other applications that rely on attention as a measure of faithfulness.

\medskip

\textbf{Ethical considerations.} While adversarial decoys could be misused to bypass defenses, we hope to encourage research toward more robust defenses and analysis tools.

\vfill
\clearpage

{\small
\section*{Acknowledgements}
This work was partially supported by project SERICS (PE00000014) under the MUR (Ministero dell'Università e della Ricerca) National Recovery and Resilience Plan funded by the European Union - NextGenerationEU.}

{
    \small
    \bibliographystyle{ieeenat_fullname}
    \bibliography{main}
}

\vfill
\clearpage

\appendix
\begin{center}
  \Large
  \textbf{Supplementary materials for the paper “Adversarial Decoys: Misdirecting Attention-Based Defenses in ViT”}
\end{center}
\vspace{1pt}
\begin{center}
\end{center}
\vspace{10pt}

\section{Additional analysis on the role of attention}
\label{app:attention_role}
We provide the derivation of Proposition~\ref{prop:attention_token_contribution} and further analyze the relationship between attention concentration and adversarial effect. The goal is to clarify why increasing the attention received by an adversarial token can amplify its influence, while also showing why attention alone is not a sufficient indicator of adversarial relevance.

\paragraph{Proof of Proposition~\ref{prop:attention_token_contribution}.}

Consider a self-attention head and a query token \(i\). The output of the head is given by
\(
    o_i = \sum_{j=0}^{n-1} A_{i,j} v_j,
\)
where \(v_j\) is the value representation of token \(j\), and
\(
    A_{i,j} =
    \frac{\exp(z_{i,j})}
    {\sum_{k=0}^{n-1}\exp(z_{i,k})}.
\)
Let \(a\) be an arbitrary token of interest. We separate the contribution of token \(a\) from the remaining tokens:
\(
    o_i = A_{i,a}v_a + \sum_{j\neq a} A_{i,j}v_j.
\)

Define
\(
    Z_{\neg a} = \sum_{k\neq a} \exp(z_{i,k})
\)
and the renormalized aggregation over all tokens except \(a\) as
\(
    \mu_{i,\neg a}
    =
    \sum_{j\neq a}
    \frac{\exp(z_{i,j})}{Z_{\neg a}} v_j.
\)
Since
\(
    A_{i,a}
    =
    \frac{\exp(z_{i,a})}{\exp(z_{i,a}) + Z_{\neg a}},
\)
the total attention assigned to all tokens different from \(a\) is
\(
    \sum_{j\neq a} A_{i,j}
    =
    1 - A_{i,a}.
\)
Moreover, for every \(j\neq a\), we can write
\[
    A_{i,j}
    =
    (1-A_{i,a})
    \frac{\exp(z_{i,j})}{Z_{\neg a}}.
\]
Therefore,
\begin{equation}
    \sum_{j\neq a} A_{i,j}v_j
    =
    (1-A_{i,a})
    \sum_{j\neq a}
    \frac{\exp(z_{i,j})}{Z_{\neg a}}v_j
    =
    (1-A_{i,a})\mu_{i,\neg a}.
\end{equation}
Substituting this expression into the attention output $ o_i$ gives the result of Proposition~\ref {prop:attention_token_contribution}
\begin{equation}
    o_i
    =
    A_{i,a}v_a
    +
    (1-A_{i,a})\mu_{i,\neg a}.
\end{equation}

\paragraph{Studying attention by constraining the optimization.}
\label{app:attention_analysis}
The previous decomposition
\(    
o_i-\mu_{i,\neg a}
    =
    A_{i,a}\left(v_a-\mu_{i,\neg a}\right)
\)
shows that the effect of token \(a\) on the output of query token \(i\) depends on two distinct factors. The first factor is the attention coefficient \(A_{i,a}\), which controls how strongly token \(a\) is weighted. The second factor is the displacement term \(v_a-\mu_{i,\neg a}\), which measures how different the value representation of token \(a\) is from the aggregate representation of the remaining tokens. This has an important consequence: attention is an amplifier, but it is not the source of the harmful direction itself. A token can receive high attention while having a value representation close to \(\mu_{i,\neg a}\), in which case its effect on the attention output remains limited. Conversely, a token can have a highly disruptive value representation, but its effect can remain weak if the corresponding attention coefficient is small.

However, despite these theoretical insights, adversarial tokens often show high attention, meaning that in practice, this is necessary to boost the effectiveness. To further study this phenomenon, we analyze whether explicitly increasing or decreasing the attention assigned to adversarial tokens affects the effectiveness of a localized attack. In particular, we consider PatchFool-style optimization and vary the strength and sign of the attention regularization term. Let \(\mathcal{P}\subseteq\mathcal{I}\) denote the set of adversarial tokens modified by the attack, and let \({L}_{\mathrm{D}}\) be the set of layers used to compute the attention objective. For a layer \(\ell\), we define the attention received by the adversarial tokens as
\begin{equation}
    S_{\mathcal{P}}^{\ell}
    =
    \frac{1}{|\mathcal{P}|}
    \sum_{p\in\mathcal{P}}
    \frac{1}{H|\mathcal{Q}|}
    \sum_{h=1}^{N_h}
    \sum_{q\in\mathcal{Q}}
    A_{q,p}^{\ell,h},
\end{equation}
where \(N_h\) is the number of attention heads, \(\mathcal{Q}\) is the set of image-token queries, and \(A_{q,p}^{\ell,h}\) is the attention assigned by query token \(q\) to adversarial token \(p\) at layer \(\ell\) and head \(h\). We then define the attention objective as
\begin{equation}
    \mathcal{L}_{\mathrm{att}}
    =
    \frac{1}{|{L}_{\mathrm{D}}|}
    \sum_{\ell\in{L}_{\mathrm{D}}}
    S_{\mathcal{P}}^{\ell}.
\end{equation}
The complete attack objective is
\(
    \max_{\delta}
    \;
    \mathcal{L}_{\mathrm{CE}}(f(x_{\delta}),y)
    +
    \lambda_{\mathrm{att}}
    \mathcal{L}_{\mathrm{att}},
\)
where \(x_{\delta}\) is the perturbed image, \(\mathcal{L}_{\mathrm{CE}}\) is the classification loss, and \(\lambda_{\mathrm{att}}\) controls the role of the attention term. Positive values of \(\lambda_{\mathrm{att}}\) encourage the adversarial tokens to receive higher attention, while negative values penalize attention concentration on the adversarial tokens. Therefore, by sweeping \(\lambda_{\mathrm{att}}\), we can evaluate whether attention is necessary for attack success or whether it mainly acts as an amplification mechanism.
Here, we use layer $5$ as in PatchFool, to compute the attention term and evaluate both the attack success rate and the attention ratio
\(
    R_{\mathcal{P}}
    =
    \frac{
    S^\ell_{\mathcal{P}}
    }{
    \max_{j\notin\mathcal{P}} S^\ell_j + \varepsilon
    },
\)
where this ratio measures whether the adversarial tokens become dominant with respect to the rest of the image.

\begin{figure*}[t]
    \centering
    \begin{subfigure}[t]{0.8\textwidth}
    \begin{subfigure}[t]{0.49\textwidth}
        \centering
        \includegraphics[width=\linewidth]{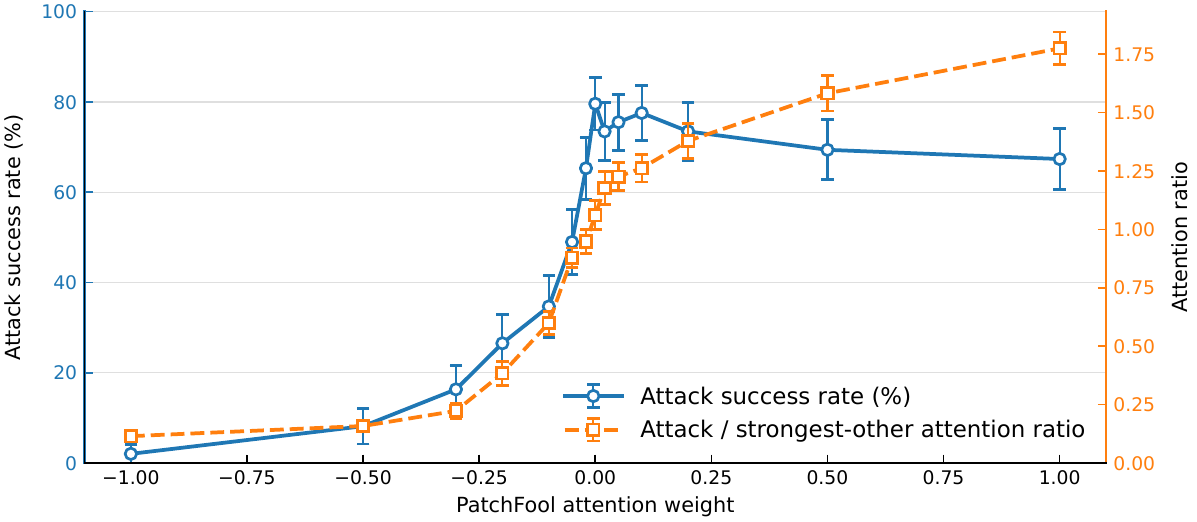}
        \caption{\small{DeiT - 1 PatchFool Token}}
    \end{subfigure}
    \hfill
    \begin{subfigure}[t]{0.49\textwidth}
        \centering
        \includegraphics[width=\linewidth]{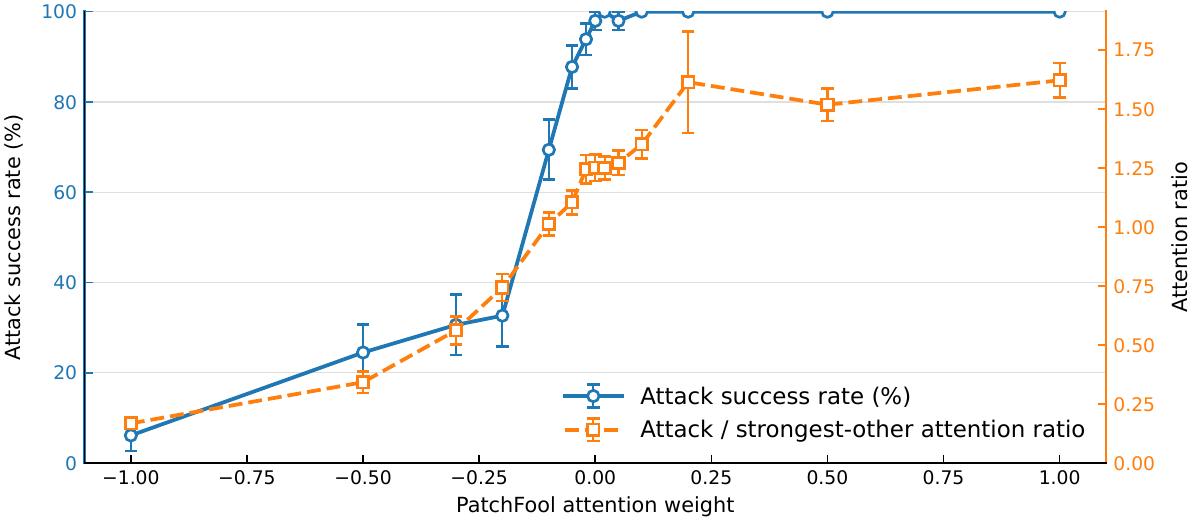}
        \caption{\small{DeiT - 2 PatchFool Tokens}}
    \end{subfigure}
    \begin{subfigure}[t]{0.49\textwidth}
        \centering
        \includegraphics[width=\linewidth]{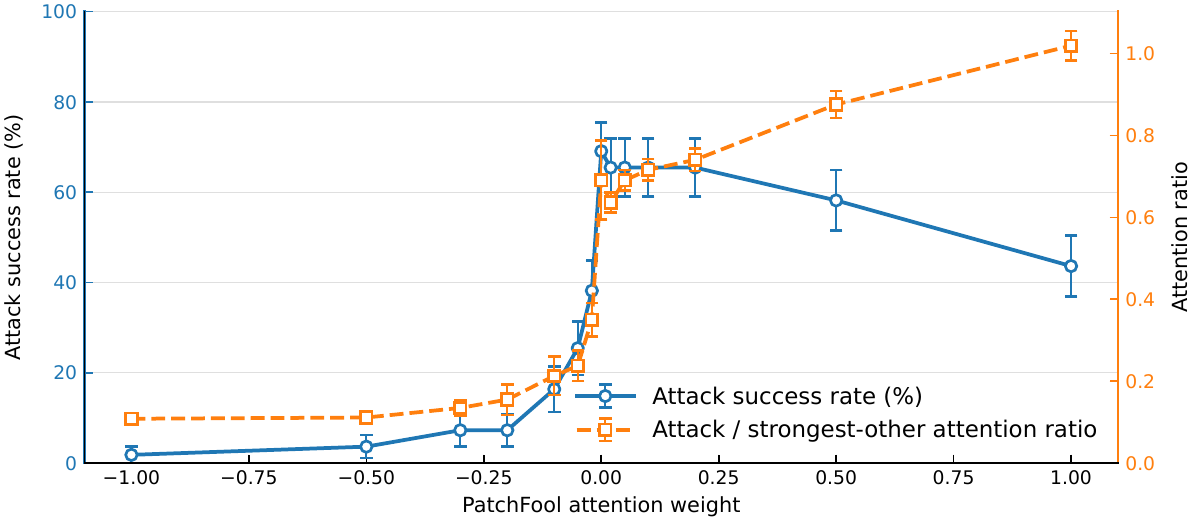}
        \caption{\small{ViT-B - 1 PatchFool Token}}
    \end{subfigure}
    \hfill
    \begin{subfigure}[t]{0.49\textwidth}
        \centering
        \includegraphics[width=\linewidth]{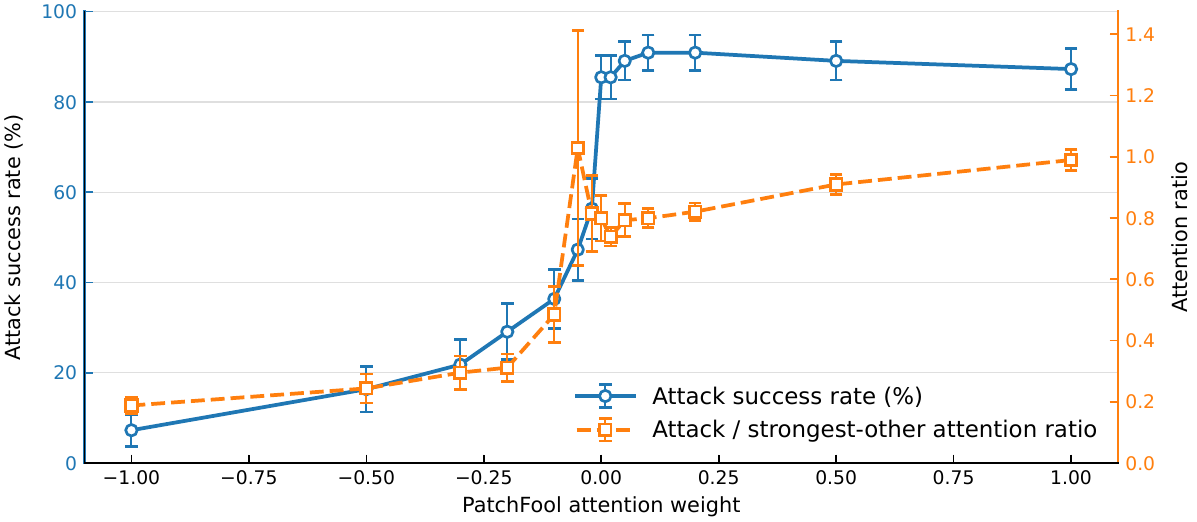}
        \caption{\small{ViT-B - 2 PatchFool Tokens}}
    \end{subfigure}
    \end{subfigure}
    \caption{\small{
        Analysis of the attack success rate and attention ratio with the attacked tokens and others across different configurations of PatchFool, for different attention weight settings at layer 5.}
    }
    \label{fig:success_analysis_patchFool}
\end{figure*}

Fig.~\ref{fig:success_analysis_patchFool} reports the results for DeiT-B and ViT-B when using one or two PatchFool tokens. When the attention weight is strongly negative, the attention ratio remains low (x axis) and the attack success rate (y axis) is also limited. As the weight approaches zero, both the attention ratio and the attack success rate increase sharply. This suggests that suppressing the attention received by adversarial tokens makes the localized attack less effective, supporting the idea that attention can amplify the adversarial contribution.

However, the curves also show that the relation is not purely monotonic. In several cases, the attack success rate reaches its maximum around small positive values of the attention weight, while larger positive values further increase the attention ratio without improving, and sometimes slightly reducing, the attack success rate. This behavior supports the interpretation given by Proposition~\ref{prop:attention_token_contribution}: increasing attention is useful only when the value displacement induced by the adversarial token remains aligned with a harmful direction for the classifier. If the optimization focuses too strongly on attention, it may produce tokens that are highly attended but not proportionally more harmful.

\begin{figure*}[t]
    \centering
    \begin{subfigure}[t]{0.75\textwidth}
    \begin{subfigure}[t]{0.49\textwidth}
    \begin{subfigure}[t]{\textwidth}
        \centering
        \includegraphics[width=\linewidth]{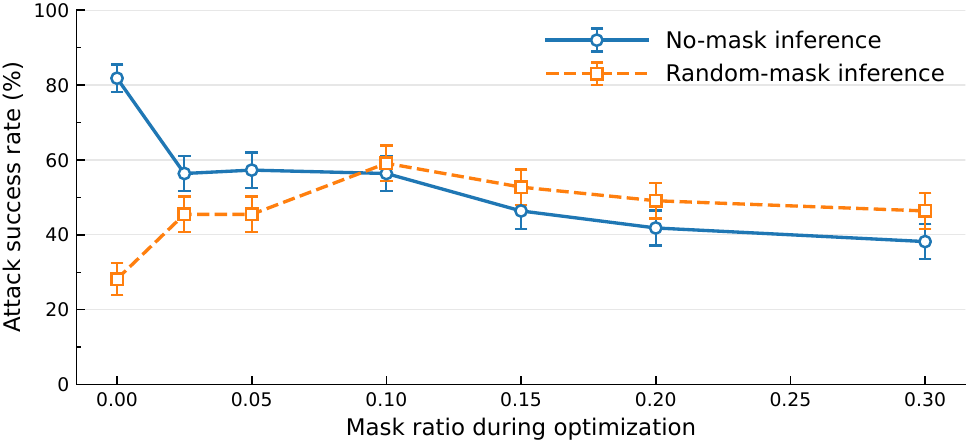}
        \caption{\small{DeiT - 1 PatchFool Token}}
    \end{subfigure}
    \hfill
    \begin{subfigure}[t]{\textwidth}
        \centering
        \includegraphics[width=\linewidth]{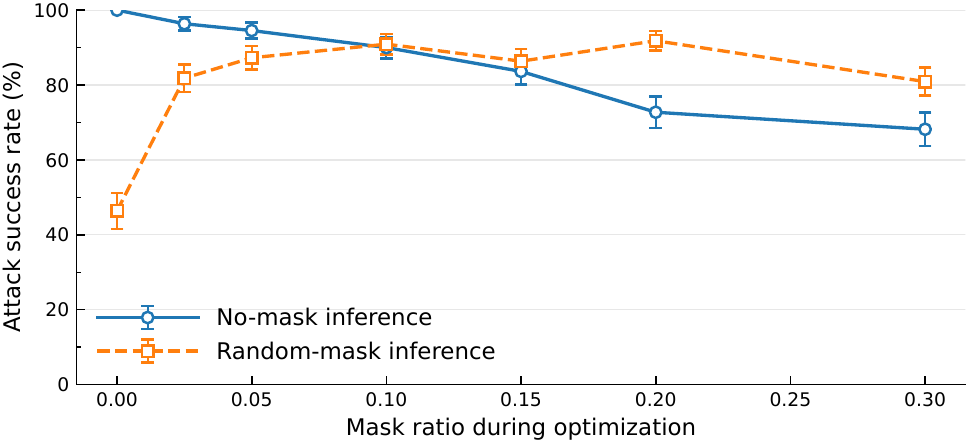}
        \caption{\small{DeiT - 2 PatchFool Tokens}}
    \end{subfigure}
    \hfill
    \begin{subfigure}[t]{\textwidth}
        \centering
        \includegraphics[width=\linewidth]{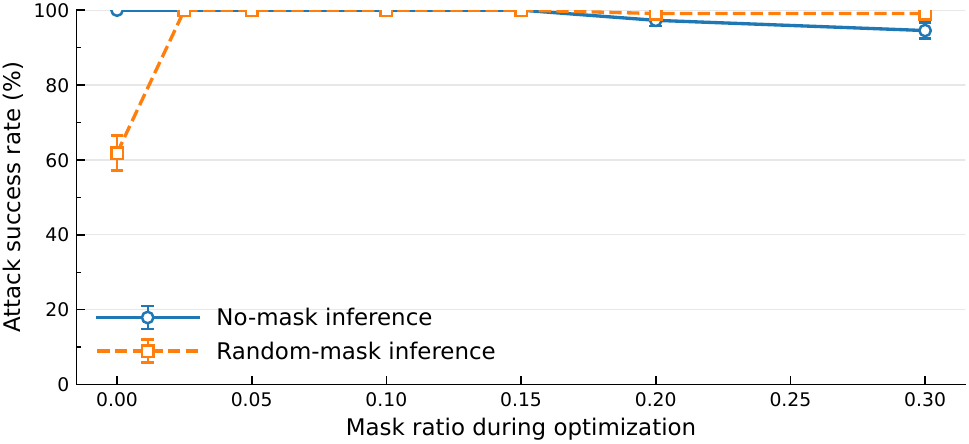}
        \caption{\small{DeiT - 4 PatchFool Tokens}}
    \end{subfigure}
    \end{subfigure}
    \begin{subfigure}[t]{0.49\textwidth}
    \begin{subfigure}[t]{\textwidth}
        \centering
        \includegraphics[width=\linewidth]{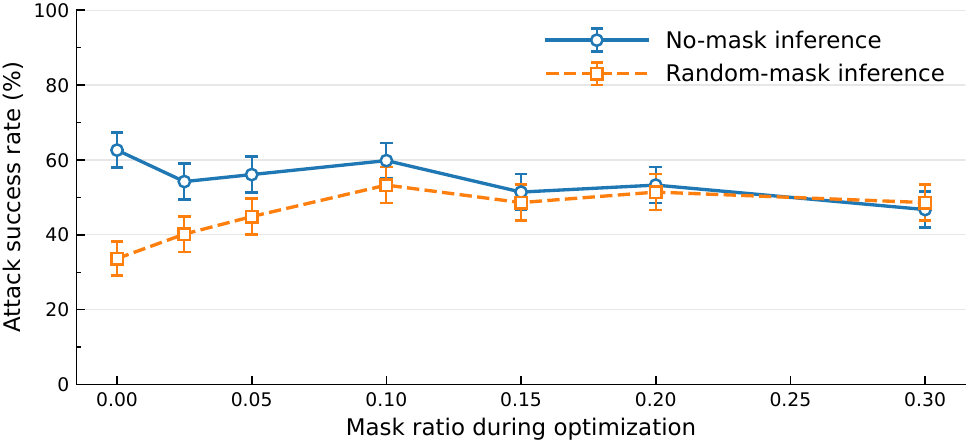}
        \caption{\small{ViT-B - 1 PatchFool Token}}
    \end{subfigure}
    \hfill
    \begin{subfigure}[t]{\textwidth}
        \centering
        \includegraphics[width=\linewidth]{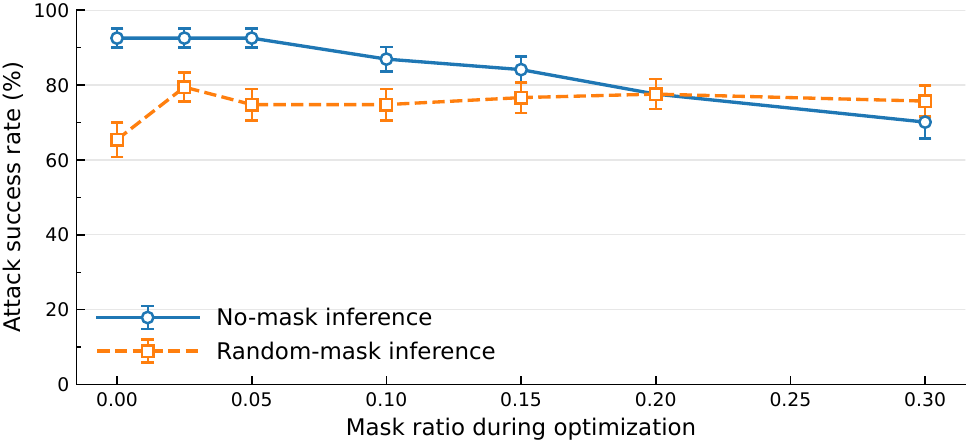}
        \caption{\small{ViT-B - 2 PatchFool Tokens}}
    \end{subfigure}
    \hfill
    \begin{subfigure}[t]{\textwidth}
        \centering
        \includegraphics[width=\linewidth]{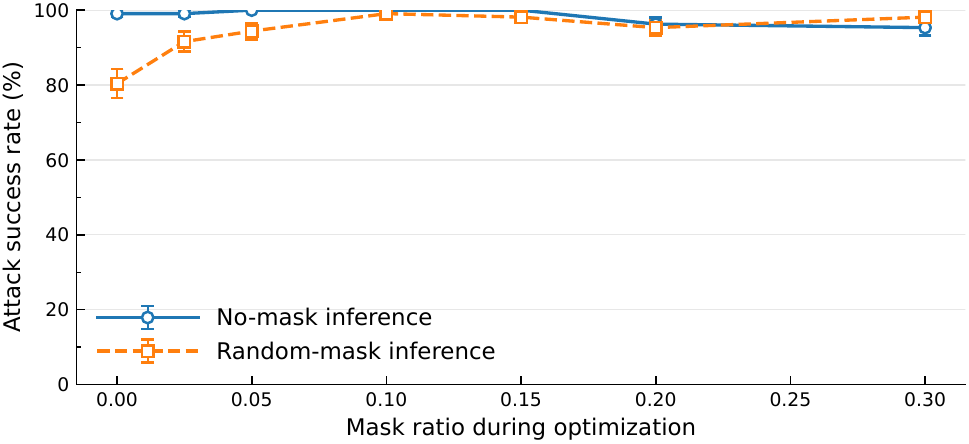}
        \caption{\small{ViT-B - 4 PatchFool Tokens}}
    \end{subfigure}
    \end{subfigure}
    \end{subfigure}
    \caption{\small{
        Effect of mask-aware optimization on localized PatchFool attacks.  The \(x\)-axis reports the mask ratio used during optimization.  Blue curves report attack success on the unmasked inference, while orange curves report attack success under random-mask inference.} 
    }
    \label{fig:mask_analysis_eot}
\end{figure*}

\section{Mask-Aware adversarial optimization.}
\label{app:mask_optimization}
As discussed in the main paper, classical localized attacks can achieve high attack success rates in a clean setting, but their effectiveness can drop significantly when random image tokens are masked, even when the masked tokens do not directly overlap with the adversarial region. This is particularly relevant for our setting: if an attention-based defense is redirected toward decoy tokens, the underlying adversarial attack remains effective only if it is sufficiently robust to the suppression of other image tokens. Therefore, to improve the robustness of localized adversarial attacks against masking-based defenses, we consider a mask-aware optimization strategy. The goal is to optimize the adversarial region under stochastic token suppression, mimicking the type of masking that may be applied at inference time. This formulation is generic and can be applied to different localized attacks, including adversarial patch attack and PatchFool.

To account for random masking during optimization, we sample at each iteration a binary random mask 
\(
    r \sim \mathcal{R}_{\rho},
\)
where \(\rho\) denotes the masking ratio. The mask \(r\in\{0,1\}^{N}\) identifies the pixels or tokens that remain visible after stochastic suppression. The randomly masked adversarial image is then
\begin{equation}
    T_r(x_{\delta})
    =
    r\odot x_{\delta}
    +
    (1-r)\odot p_{\mathrm{mask}},
\end{equation}
where \(p_{\mathrm{mask}}\) is the replacement value used for masked regions, such as zero, gray, or the image mean.

The mask-aware objective optimizes the perturbation over the expectation induced by the masking distribution:
\begin{equation}
    \max_{\delta}
    \;
    \mathbb{E}_{r\sim\mathcal{R}_{\rho}}
    \left[
        \mathcal{L}_{\mathrm{adv}}
        \left(
            f(T_r(x_{\delta})), y
        \right)
    \right],
    \label{eq:random_mask_eot}
\end{equation}
where \(\mathcal{L}_{\mathrm{adv}}\) is the adversarial objective. 
This formulation can be interpreted as a robustness-oriented optimization of the localized attack. Instead of optimizing the perturbation only for the original image, the attack is optimized over a distribution of partially masked images. As a result, the perturbation is encouraged to remain effective even when some tokens are removed or replaced. 

Fig.~\ref{fig:mask_analysis_eot} reports the effect of mask-aware optimization for PatchFool attacks with different token budgets. The \(x\)-axis denotes the masking ratio used during optimization, while the curves report the attack success rate in the undefended setting and under random-mask inference with a masking ratio of \(10\%\). The results show that introducing a moderate masking ratio during optimization generally improves the effectiveness of the attack under random-mask inference. This effect is particularly evident when multiple PatchFool tokens are used. For example, with two or four tokens, the attack success rate under random-mask inference increases substantially compared to the case where no random masking is used during optimization. The behavior also reveals a trade-off. When the masking ratio is too small, the attack is optimized mainly for the unmasked image and can be less robust to stochastic suppression at inference time. Conversely, when the masking ratio becomes too large, the optimization may become overly conservative or less aligned with the original undefended objective, leading to a reduction in attack success without defense. Therefore, moderate masking ratios already provide a better compromise between robustness to random masking and standard attack effectiveness. Based on this analysis, in the main experiments, we use a token-level masking probability of \(0.05\) during attack optimization (excluding tokens associated with the adversarial attacks).

\section{Additional Results}
\label{appx:additional_results}

In this section, we report additional analyses supporting the paper. Note that in this section layers are indexed from $0$ to $L-1$, rather than from $1$ to $L$ as in the main paper.

\subsection{Results with top-$k$ analysis}
\label{appx:top_k_analysis}
\textbf{Number of decoy ablation.} Tab.~\ref{tab:topk_ndecoy_abl} reports the results of the top-$k$ analysis under different numbers of decoy tokens ($1$, $2$, and $4$). It can be noticed, in bold, that increasing the number of tokens leads to slightly improved coverage in deep layers. These results are consistent with the results extrapolated in the defense setting, where such improvement translates into more effective steering behavior.

\begin{table}[t]
\centering
\footnotesize
\setlength{\tabcolsep}{3pt}
\resizebox{\columnwidth}{!}{%
\begin{tabular}{cc|cccccccccccc}
\hline
\# Patches & Model & L0 & L1 & L2 & L3 & L4 & L5 & L6 & L7 & L8 & L9 & L10 & L11 \\
\hline

\multirow{3}{*}{1}
& DeiT & \textbf{98} & \textbf{71} & \textbf{82} & \textbf{100} & \textbf{100} & \textbf{100} & 97 & 91 & 74 & 75 & 53 & 99 \\
& ViT-B & \textbf{70} & \textbf{95} & \textbf{91} & 99 & 87 & 59 & 17 & 24 & 1 & 2 & 19 & 95 \\
& ViT-S & \textbf{100} & \textbf{100} & \textbf{100} & \textbf{100} & \textbf{100} & \textbf{100} & 68 & 16 & 19 & 30 & 48 & 93 \\
\hline

\multirow{3}{*}{2}
& DeiT & 96 & 49 & 76 & \textbf{100} & \textbf{100} & \textbf{100} & \textbf{98} & \textbf{96} & 75 & 79 & 73 & \textbf{100} \\
& ViT-B & 61 & 94 & 89 & \textbf{100} & 94 & 77 & 32 & 37 & 3 & 3 & 27 & 98 \\
& ViT-S & 97 & \textbf{100} & \textbf{98} & \textbf{100} & \textbf{100} & \textbf{100} & 82 & 21 & 22 & 32 & 51 & 95 \\
\hline

\multirow{3}{*}{4}
& DeiT & 89 & 42 & 74 & \textbf{100} & \textbf{100} & \textbf{100} & \textbf{98} & \textbf{96} & \textbf{86} & \textbf{91} & \textbf{96} & \textbf{100} \\
& ViT-B & 29 & 85 & 89 & \textbf{100} & \textbf{96} & \textbf{85} & \textbf{49} & \textbf{52} & \textbf{17} & \textbf{18} & \textbf{35} & \textbf{99} \\
& ViT-S & 88 & 97 & 93 & \textbf{100} & \textbf{100} & \textbf{100} & \textbf{96} & \textbf{43} & \textbf{42} & \textbf{46} & \textbf{64} & \textbf{99} \\
\hline
\end{tabular}
}
\caption{\small{Results of the top-$k$ overlap (\%) analysis across layers for $1$, $2$, and $4$ optimized decoy patches. Bold indicates the highest value for the same model across patch counts.}}
\label{tab:topk_ndecoy_abl}
\end{table}

\textbf{All vs single layer.} Tab.~\ref{tab:top4_allvssingle} extends the top-$k$ analysis by comparing joint optimization across all layers with optimization performed on each layer individually. In this case, experiments are conducted using $4$ adversarial decoy tokens on the three ViT architectures considered in this work.
As expected, optimizing each layer independently leads to higher top-$k$ coverage analyzed on the same layer, since the optimization no longer needs to satisfy competing objectives across different depths of the architectures. However, an interesting trend can be noticed: the layers that are difficult to optimize in the joint setting remain the most challenging even when optimized in isolation. This behavior is particularly evident for ViT-B and ViT-S, where several intermediate layers consistently exhibit lower decoy coverage in both settings. These results suggest that promoting the decoy tokens to the top of the attention ranking is intrinsically more difficult at specific depths of the architecture, rather than being a consequence of joint optimization across layers. Therefore, simply focusing on the most challenging layers does not eliminate their inherent optimization difficulty. That said, although informative for understanding the optimization dynamics, single-layer optimization is of limited practical interest for bypassing attention-based defenses. Indeed, defenses typically exploit attention patterns across multiple layers, and optimizing decoys for only one layer does not produce a sufficiently consistent shift throughout the overall network.

\begin{table}[t]
\centering
\footnotesize
\setlength{\tabcolsep}{3pt}
\resizebox{\columnwidth}{!}{%
\begin{tabular}{c|cccccccccccc}
\hline
Model & L0 & L1 & L2 & L3 & L4 & L5 & L6 & L7 & L8 & L9 & L10 & L11 \\
\hline

DeiT all & 89 & 42 & 74 & 100 & 100 & 100 & 98 & 96 & 86 & 91 & 96 & 100 \\
DeiT sep & 100 & 100 & 100 & 100 & 100 & 100 & 100 & 100 & 100 & 100 & 100 & 100 \\
\hline

ViT-B all & 29 & 85 & 89 & 100 & 96 & 85 & 49 & 52 & 17 & 18 & 35 & 99 \\
ViT-B sep & 100 & 100 & 100 & 100 & 100 & 100 & 97 & 90 & 67 & 50 & 84 & 100 \\
\hline

ViT-S all & 88 & 97 & 93 & 100 & 100 & 100 & 96 & 43 & 42 & 46 & 64 & 99 \\
ViT-S sep & 100 & 100 & 100 & 100 & 100 & 100 & 100 & 100 & 98 & 94 & 96 & 100 \\
\hline
\end{tabular}
}
\caption{\small{Top-$k$ overlap (\%) per layer obtained by optimizing $4$ decoys independently for each layer, for DeiT, ViT-B, and ViT-S.}}
\label{tab:top4_allvssingle}

\end{table}

\begin{figure*}[t]
    \centering
    \begin{minipage}[t]{0.45\textwidth}
        \centering
        \includegraphics[width=\linewidth]{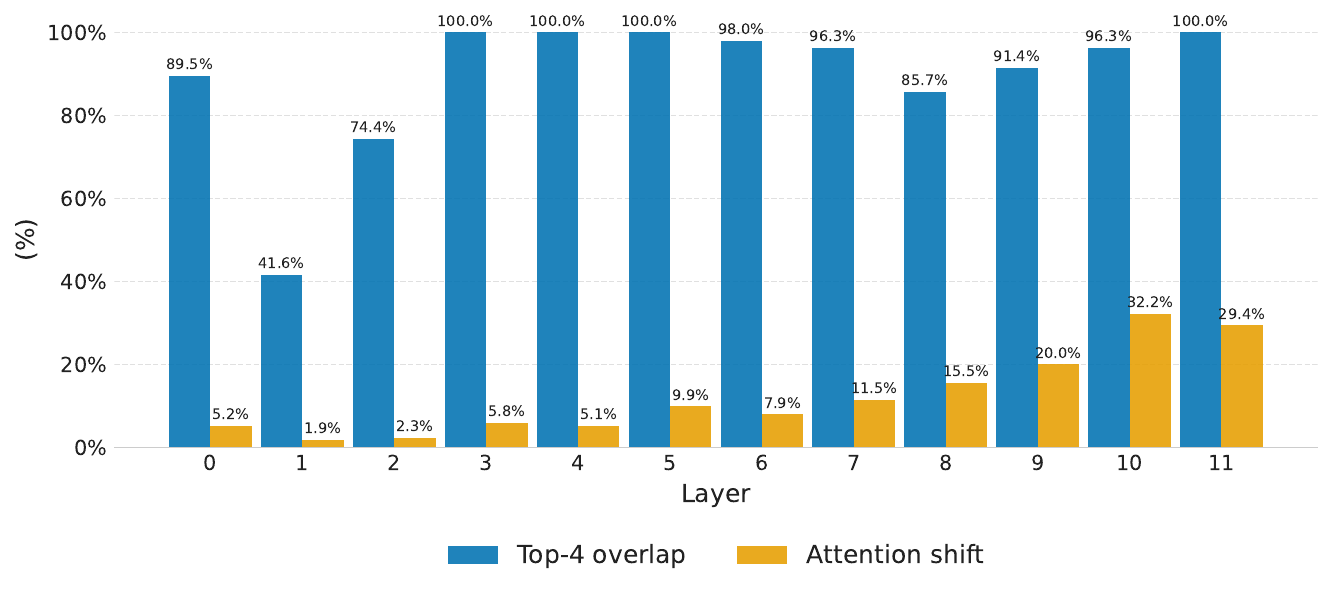}
        \subcaption{DeiT -- optimization all-layers}
        \label{fig:deit_alllayers}
    \end{minipage}
    \hfill
    \begin{minipage}[t]{0.45\textwidth}
        \centering
        \includegraphics[width=\linewidth]{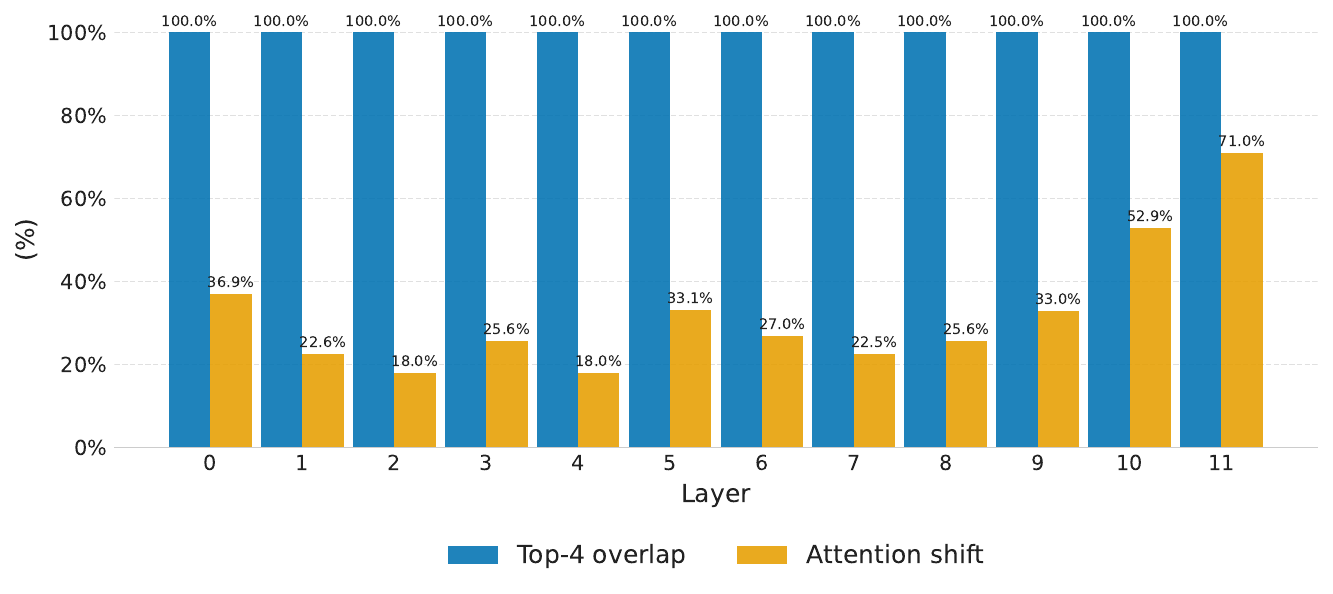}
        \subcaption{DeiT -- optimization single-layer}
        \label{fig:deit_singlelayers}
    \end{minipage}

    \vspace{0.8em}

    \begin{minipage}[t]{0.45\textwidth}
        \centering
        \includegraphics[width=\linewidth]{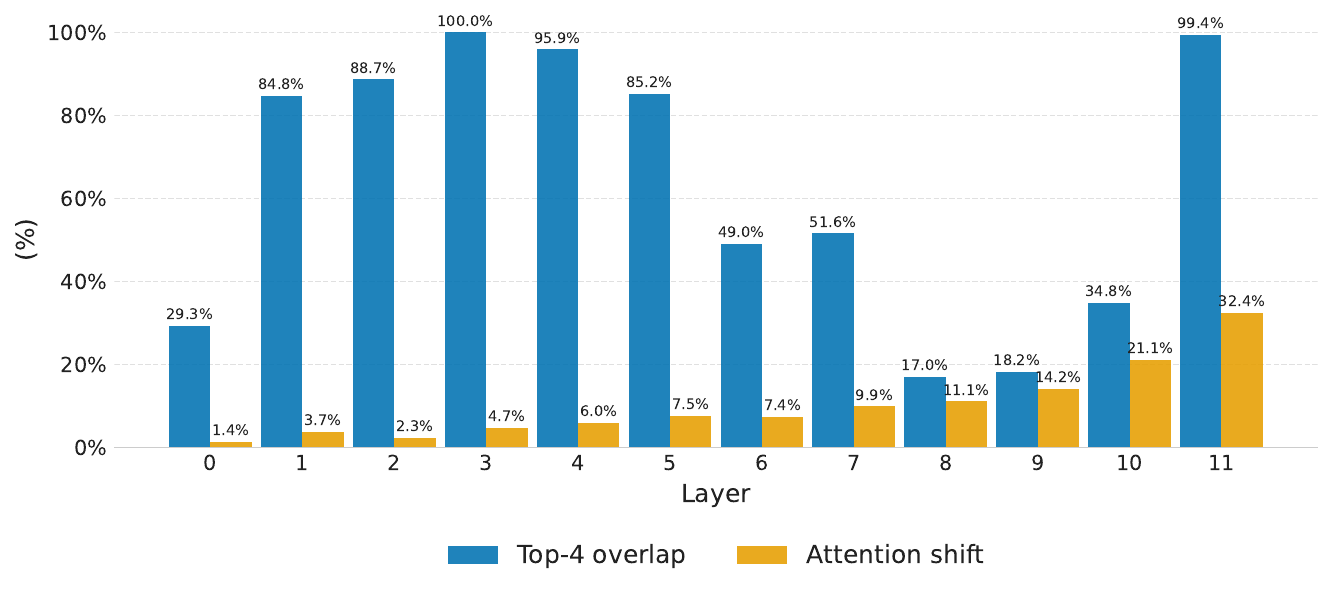}
        \subcaption{ViT-B -- optimization all-layers}
        \label{fig:vit_alllayers}
    \end{minipage}
    \hfill
    \begin{minipage}[t]{0.45\textwidth}
        \centering
        \includegraphics[width=\linewidth]{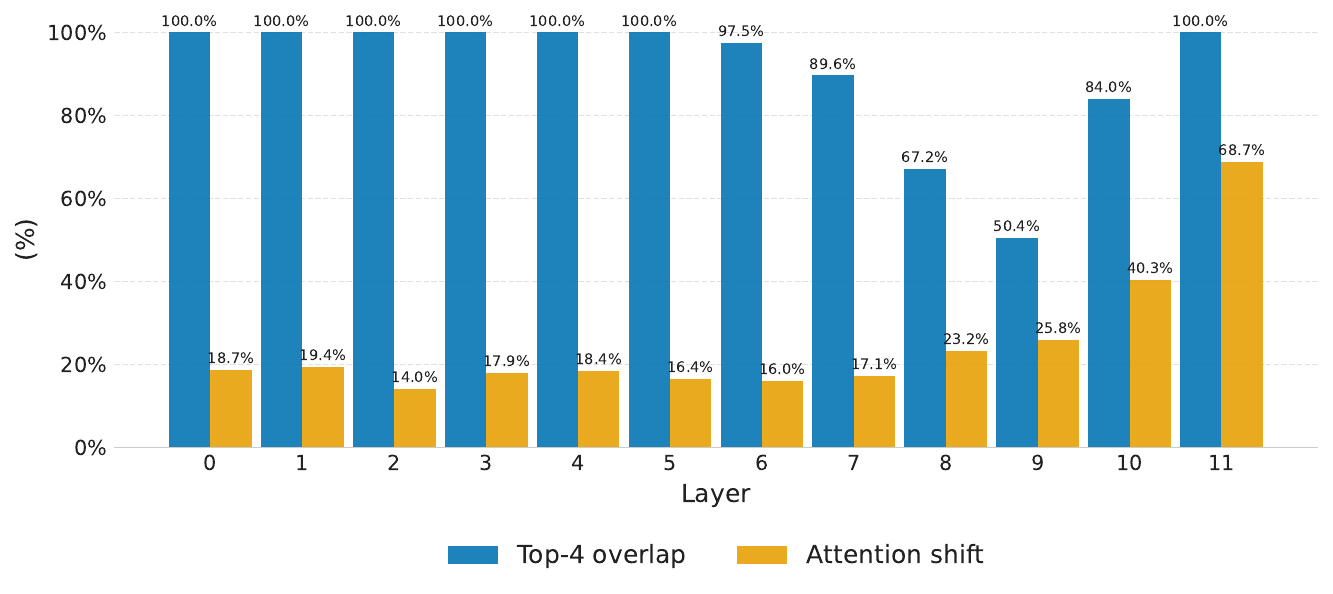}
        \subcaption{ViT-B -- optimization single-layer}
        \label{fig:vit_singlelayers}
    \end{minipage}

    \caption{Top-$4$ overlap and attention shift trend comparison on target tokens across layers. The top row shows DeiT, while the bottom row shows ViT-B. On the left, results are computed by optimizing all layers together, while on the right, single-layer optimization is performed.}
    \label{fig:topk_shift_analysis}
\end{figure*}

The top-$k$ analysis, however, only highlights whether the targeted tokens manage to appear among the highest-ranked ones. Understanding how much attention is actually shifted toward these tokens can also be insightful. Fig.~\ref{fig:topk_shift_analysis} compares the top-$k$ analysis with the attention-shift metric, computed as follows: we first normalize the attention distributions after excluding the CLS token, and then measure the difference in the total attention mass assigned to the target tokens before and after applying the decoy tokens.
It is worth noting that, due to the softmax normalization, attention is redistributed across tokens rather than accumulated independently on a given subset. As a result, even strong shifts in attention toward the target tokens remain bounded and do not necessarily reach saturation. Nevertheless, meaningful patterns can still be observed.
In particular, while the top-$k$ analysis shows that promoting target tokens to the highest-attended positions in deeper layers can be challenging, the shift metric reveals a consistent increase across layers, especially in the DeiT model. This suggests that, even when target tokens do not enter the top-$k$ set, attention is still progressively redistributed toward them as the network depth increases.

\subsection{{Optimization details, iterations and costs}}
\label{appx:optimization_details}
\begin{figure}[h]
    \centering
    \begin{subfigure}[t]{0.45\textwidth}
    \includegraphics[width=\textwidth]{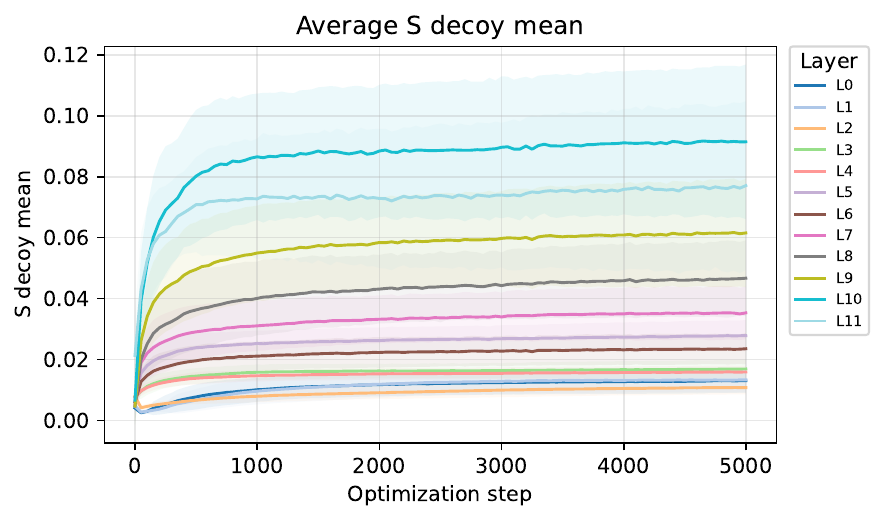}
    \end{subfigure}
    \begin{subfigure}[t]{0.45\textwidth}
    \includegraphics[width=\textwidth]{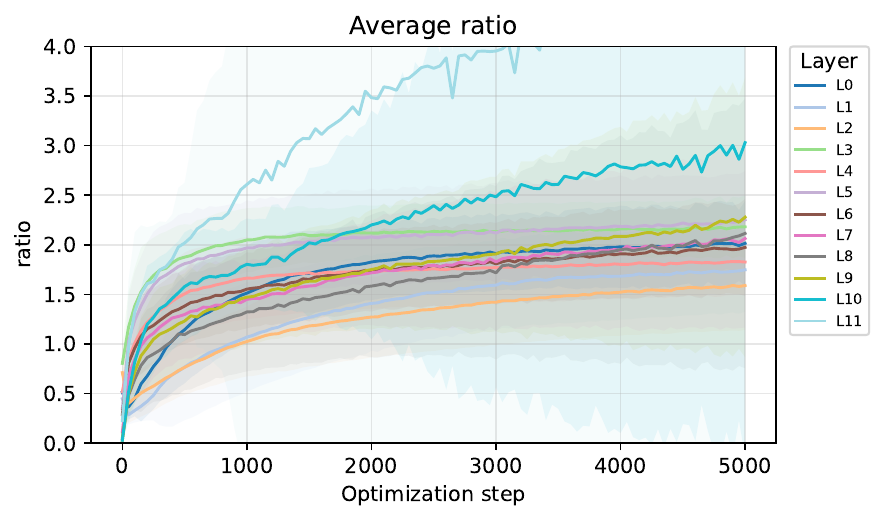}
    \end{subfigure}
    \vspace{-1em}
    \caption{\small{Evolution of decoy optimization across layers. The plots show the average decoy attention score (top) and decoy-to-competing-token ratio (bottom) over optimization steps, with shaded regions indicating variability across images.}}
    \label{fig:loss_analysis}
\end{figure}

Regarding the number of steps required to optimize the decoys, we acknowledge that jointly optimizing a layer-wise objective across multiple transformer layers is a challenging process. For stability, we therefore use a relatively low learning rate, set to \rev{$0.05$}, which requires a larger number of optimization steps to obtain effective decoy behavior. Fig.~\ref{fig:loss_analysis} reports the average evolution of the decoy attention score and the corresponding ratio across layers during optimization. The plots show that most of the improvement occurs in the early phase, while the curves continue to increase more gradually as the number of steps grows, suggesting that longer optimization can further strengthen the decoy effect. 
In the main experiments, we use 2500 optimization steps as a trade-off between computational cost and decoy effectiveness. Nevertheless, these results indicate that stronger performance could potentially be achieved by increasing the number of optimization steps, especially when computational resources are less constrained. 

\begin{figure*}[t]
    \centering
    \includegraphics[width=0.9\linewidth]{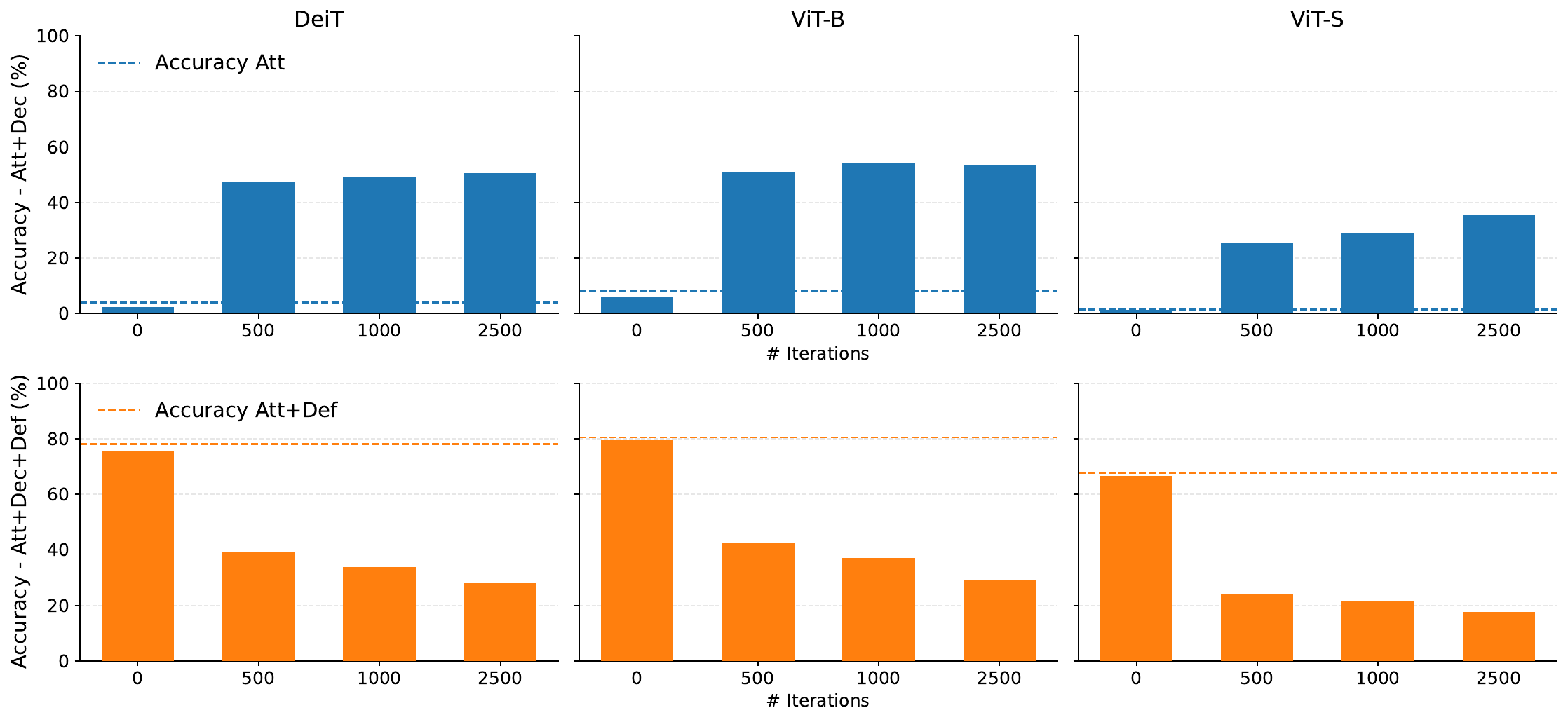}
    \vspace{-1em}
    \caption{\small{\rev{Analysis of the effect of $0$ (random decoys), $500$, $1000$, and $2500$ iterations for the decoys optimization. On top, the accuracy is computed on the attacked images with decoys before the defense. On the bottom the accuracy is computed on the attacked images with decoys after the defense. In all experiments, $4$ decoys are considered, and the PatchFool mask-aware attack is employed with $2$ patches. The dashed lines represent the baseline accuracy obtained with the adversarial patches alone (without decoys), before (blue) and after (orange) the defense is applied, allowing direct comparison against the decoy-based results.}}}
    \label{fig:iter_study}
\end{figure*}

\rev{\paragraph{Optimization impact on defended accuracy}}
\rev{To verify that the trend observed in the optimization objective translates into improved effectiveness against the defense, we additionally evaluate decoys optimized for different numbers of steps: $0$ (which corresponds to the completely unoptimized case, i.e., random values), $500$, $1000$ and $2500$. As shown in Fig.~\ref{fig:iter_study}, increasing the number of optimization steps makes the decoys progressively more effective at evading the defense (bottom row), although, in contrast, the attack in the absence of defenses becomes less effective as the decoys are further optimized. This behavior is expected, as when the decoys are randomly initialized, they do not consistently alter the adversarial attention pattern, whereas optimization progressively encourages them to attract attention away from the adversarial attack, weakening it but enhancing the defense bypass. These results, together with the one presented in the main paper (Tab.~\ref{tab:main_results}), and with the analysis in App.~\ref{appx:adverarial_effect_decoy}, provide a direct confirmation that the defense misdirection is driven by the decoy optimization rather than just adding randomly initialized patches.}

\rev{Importantly, in the comparative analyses with adversarial attacks (e.g., Sec.~\ref{sec:budget_analysis} and App.~\ref{appx:adaptive_attack}), we do not claim that decoy optimization is computationally more efficient than adversarial optimization. Indeed, adversarial patches typically converge in fewer optimization steps (e.g., $250$ or $500$) than those required to obtain effective decoy behavior. Rather, our goal is to investigate a different attack paradigm, in which part of the attack patch budget is allocated to decoys instead of optimizing all injected components adversarially. Our results show that such a hybrid strategy can provide stronger attack performance against the considered defenses, highlighting a trade-off between optimization cost and attack effectiveness.}

\begin{figure}[t]
    \centering
    \begin{subfigure}[t]{0.48\textwidth}
    \includegraphics[width=\linewidth]{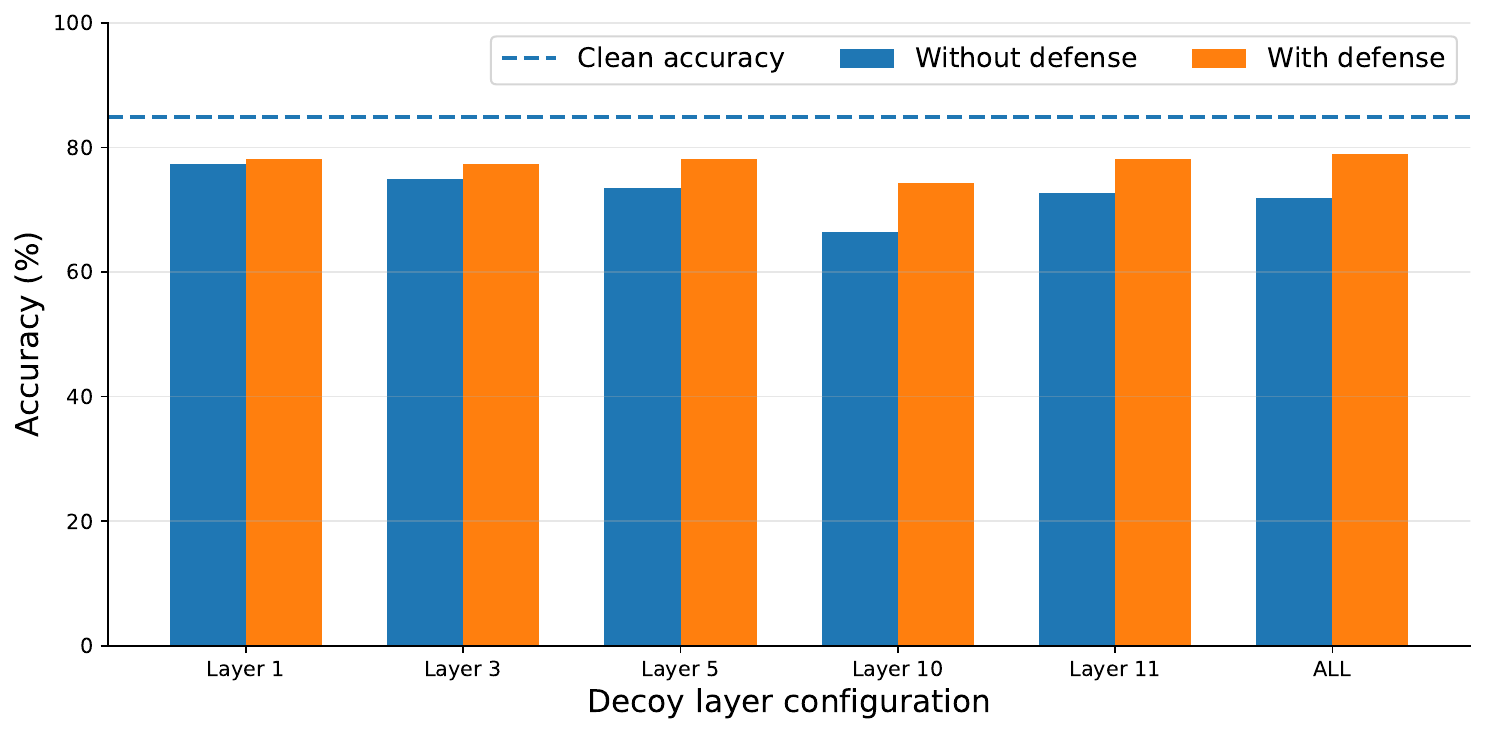}
    \caption{\small{DeiT}}
    \end{subfigure}
    \begin{subfigure}[t]{0.48\textwidth}
    \includegraphics[width=\linewidth]{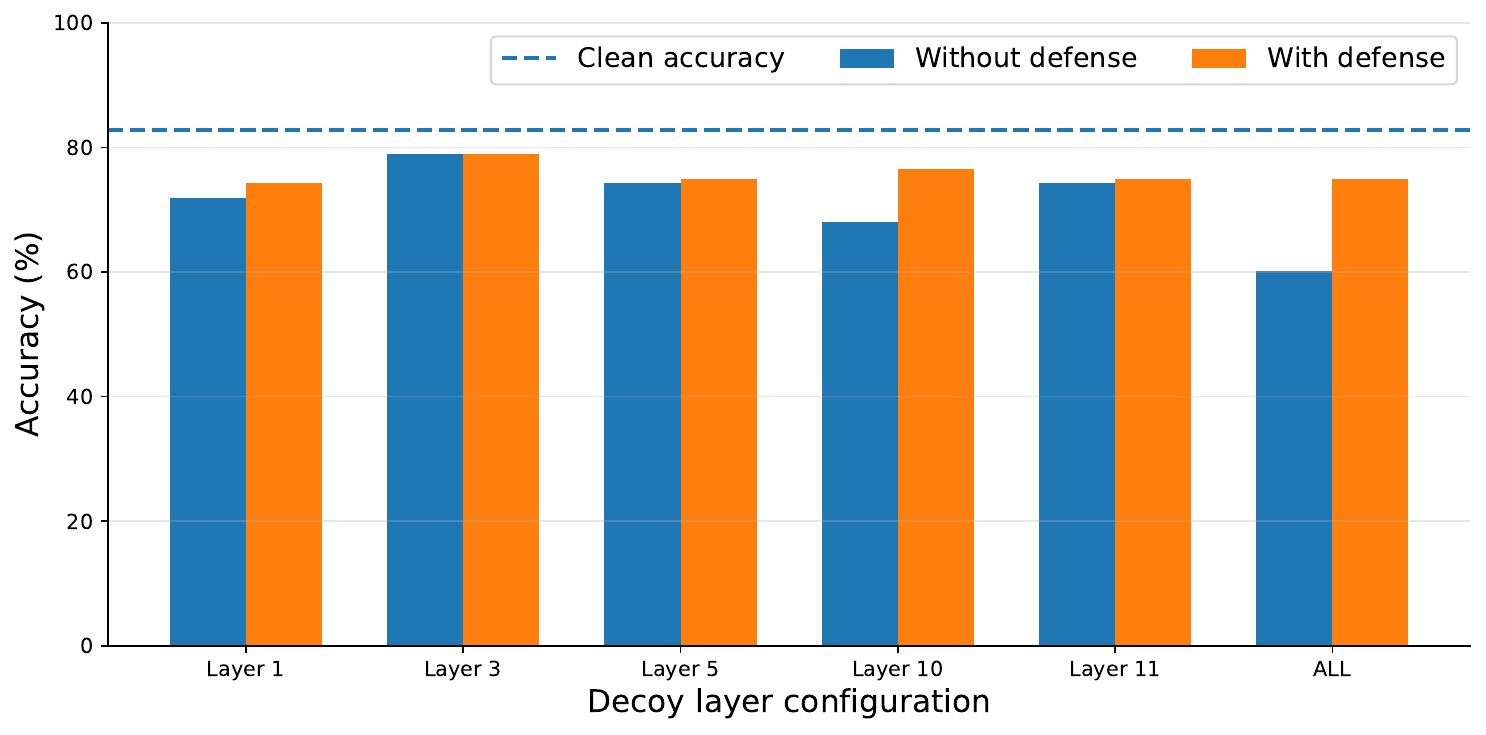}
    \caption{\small{ViT-S}}
    \end{subfigure}
    \vspace{-1em}
    \caption{\rev{\small{Classification accuracy obtained when applying only the optimized decoys, without any adversarial attack, for DeiT (top) and ViT-S (bottom). Decoys are optimized either with respect to individual transformer layers or jointly across all layers (\emph{ALL}), and evaluated without defense (blue) and with ARMRO (orange). The dotted lines report the clean-model accuracy.}}}
    \label{fig:decoy_adversarial_analysis}
\end{figure}

\rev{
\subsection{On the adversarial effect of decoys}
\label{appx:adverarial_effect_decoy}
We further investigate whether the decoy optimization inherently induces an adversarial effect, i.e., whether optimizing a patch to attract model attention also directly causes misclassification. To isolate this effect, we evaluate classification accuracy when applying only the optimized decoys, both with and without the defense mechanisms, without introducing any adversarial attack. This setting is intentionally different from the main experiments, where decoys are used to misdirect the defense and thereby preserve the effectiveness of the actual adversarial attacks.

As shown in Fig.~\ref{fig:decoy_adversarial_analysis}, for DeiT (top) and ViT-S (bottom), the decoys alone induce only a limited degradation in classification accuracy. We optimize the decoys under different configurations, either jointly across all transformer layers, as done in the main experiments, to more effectively misdirect the defense, or with respect to individual layers, as reported along the $x$-axis. Decoys optimized at deeper layers tend to produce a slightly larger, although still limited, decrease in accuracy compared with those optimized at shallower layers. This behavior is expected, as perturbations affecting representations closer to the final classification layers can have a stronger influence on the model decision.
These results indicate that the optimized decoys do not exhibit a substantial adversarial effect on their own. Rather, their primary role is to manipulate the attention patterns exploited by the defense, thereby misdirecting its intervention and allowing the actual adversarial perturbation to retain its effectiveness, as demonstrated in the main experiments.
}

\subsection{Analysis with other defenses}
\label{appx:additional_defense}
\rev{In addition to the results obtained with ARMRO, which is considered in the main paper as a state-of-the-art attention-based defense against token-level attacks, we further investigate whether the proposed decoy strategy remains effective against alternative attention-suppression mechanisms.}

\rev{
\paragraph{Analysis of top-\(K\) suppression.}
To extend suppression-based approaches previously explored in the CNN literature against adversarial patch attacks \cite{rossolini2023defending, urtasun_NIPS2016_c8067ad1, yu2021defending}, we consider a layer-wise masking defense based on the suppression of the most attended tokens. Specifically, we evaluate the mask-aware PatchFool attack in Tab.~\ref{tab:supp_topk_layerwise_masking} using \(1\), \(2\), and \(4\) adversarial tokens, as in the main paper, while keeping the number of decoy tokens fixed to \(4\).

Unlike ARMRO, which applies a more selective masking strategy, this baseline independently identifies the top-\(3\) most attended tokens at each transformer layer and suppresses the union of the selected tokens. Consequently, the total number of suppressed tokens can be substantially larger than with ARMRO, depending on the overlap among the tokens selected across different layers. Such a defense, while simpler than ARMRO, can often provide greater robustness at the cost of a larger number of suppressed tokens and therefore a lower clean-defended accuracy, resulting in a less favorable robustness-accuracy trade-off. To the best of our knowledge, ARMRO represents a state-of-the-art approach in this regard.

As shown in Tab.~\ref{tab:supp_topk_layerwise_masking}, the results are consistent with those obtained in the main paper. In particular, combining decoy and adversarial tokens consistently increases the adversarial effect in the presence of the defense (\emph{Att+Dec+Def}) across all considered adversarial-token budgets. These results indicate that the effectiveness of the proposed decoy mechanism is not specific to the masking strategy adopted by ARMRO, but also extends to a more aggressive layer-wise top-\(K\) suppression scheme. This further supports the general principle underlying our approach: by attracting the attention-based suppression mechanism toward the decoy tokens, the actual adversarial tokens are more likely to remain unaffected by the defense.

\begin{table}[t]
\centering
\scriptsize
\setlength{\tabcolsep}{2.5pt}
\renewcommand{\arraystretch}{1.15}

\newcommand{\acc}[2]{%
    \ensuremath{#1{\scriptscriptstyle\,\pm\,#2}}%
}

\newcommand{\rowstrut}{\rule{0pt}{2.5ex}}

\newcommand{\ResultPair}[5]{%
    #1
        & \rowstrut #2
        & #3
        & #4
        & #5 \\
}

\resizebox{0.48\textwidth}{!}{%
\begin{tabular}{@{}c|c|c|c|c|c|c@{}}

\midrule

\shortstack{Model} &
\shortstack{\#Adv.T} &
\shortstack{Clean} &
\shortstack{Att} &
\shortstack{Att+Dec} &
\shortstack{Att+Def} &
\shortstack{Att+Dec+Def} \\

\midrule


\multirow{3}{*}{\textbf{DeiT}}
& 1
& 82.32
& \acc{36.82}{1.51}
& \acc{61.82}{1.52}
& \acc{80.96}{1.23}
& \acc{39.84}{1.53} \\

& 2
& -
& \acc{6.84}{0.79}
& \acc{48.34}{1.56}
& \acc{80.47}{1.24}
& \acc{12.40}{1.03} \\

& 4
& -
& \acc{0.20}{0.14}
& \acc{34.67}{1.49}
& \acc{76.86}{1.32}
& \acc{1.95}{0.43} \\

\midrule


\multirow{3}{*}{\textbf{ViT-B}}
& 1
& 85.06
& \acc{41.50}{1.54}
& \acc{65.04}{1.49}
& \acc{82.81}{1.18}
& \acc{48.93}{1.56} \\

& 2 
& -
& \acc{12.70}{1.04}
& \acc{54.49}{1.56}
& \acc{81.15}{1.22}
& \acc{24.32}{1.34} \\

& 4
& -
& \acc{0.68}{0.26}
& \acc{35.25}{1.49}
& \acc{74.80}{1.36}
& \acc{9.18}{0.90} \\

\midrule


\multirow{3}{*}{\textbf{ViT-S}}
& 1
& 81.64
& \acc{18.85}{1.22}
& \acc{50.49}{1.56}
& \acc{78.32}{1.29}
& \acc{24.61}{1.35} \\

& 2 
& -
& \acc{3.12}{0.54}
& \acc{35.94}{1.50}
& \acc{77.54}{1.30}
& \acc{11.43}{0.99} \\

& 4
& -
& \acc{0.20}{0.14}
& \acc{12.11}{1.02}
& \acc{74.22}{1.37}
& \acc{5.57}{0.72} \\

\bottomrule
\end{tabular}%
}
\caption{\small{
\rev{
Results under the PatchFool mask-aware attack on $1024$ ImageNet-Val samples, considering different numbers of adversarial tokens and a top-$3$ layer-wise masking defense. The defense suppresses the union of the three most-attended tokens selected at each transformer layer. All configurations employ $4$ decoy tokens, as in the main experiments in Sec.~\ref{sec:main_results}.
}
}}
\label{tab:supp_topk_layerwise_masking}
\end{table}

\paragraph{Analysis of RSA against Decoy.}
For completeness, we also evaluate our decoy patches against the RSA defense~\cite{mu2021defending}. RSA is an early defense for ViTs against adversarial patch attacks and differs from ARMRO in both its assumptions and its defense mechanism. In particular, RSA is designed for a single square patch of known size, selects tokens based on attention values rather than attention weights, and performs runtime masking within the attention matrices rather than masking the original image.

To provide a comparison under RSA, we evaluate our method against PatchFool and the adversarial patch attack. The results are reported in Tab.~\ref{tab:rsa_results}. For both attacks, we consider \(1\) and \(4\) attack patches while keeping the number of decoy tokens fixed to \(4\). In all experiments, we use a defense window of \(4\), i.e., the window size used by the selection criterion and, consequently, the number of tokens that can be suppressed within each layer.
The \emph{Att+Def} results already highlight an important difference between RSA and ARMRO: RSA is effective when the attack configuration closely matches the assumptions of its defense mechanism (on adversarial patches where attack patches form a square), and when the window size corresponds to the number of patches in the attacked region. RSA effectiveness decreases as the attack configuration deviates from this setting, and for this reason, results are less readable. However, we can notice that decoys provide a more limited benefit under RSA than under ARMRO. We hypothesize that this behavior is first related to the fact that adding decoys introduces additional tokens beyond the single-patch setting considered by RSA, which makes the resulting behavior less clear. However, this behavior probably further stems from two properties of the defense mechanism. First, its selection criterion is based on attention values rather than attention weights and is therefore less directly affected by the attention redistribution induced by our decoys. Second, RSA performs masking during the forward pass rather than as a final masking step. Consequently, tokens affected by the decoys may be suppressed at an early layer, while the adversarial patch can remain identifiable in subsequent layers, reducing the ability of the decoys to consistently misdirect the defense throughout the network.
}

\begin{table}[t]
\centering
\scriptsize
\setlength{\tabcolsep}{2.5pt}
\renewcommand{\arraystretch}{1.15}

\newcommand{\acc}[2]{%
    \ensuremath{#1{\scriptscriptstyle\,\pm\,#2}}%
}

\newcommand{\rowstrut}{\rule{0pt}{2.5ex}}

\newcommand{\ResultPair}[5]{%
    #1
        & \rowstrut #2
        & #3
        & #4
        & #5 \\
}

\resizebox{0.48\textwidth}{!}{%
\begin{tabular}{@{}c|c|c|c|c|c|c@{}}

\midrule
\multicolumn{7}{c}{%
    PatchFool~\cite{fu2022patch}
} \\
\midrule

\shortstack{Model} &
\shortstack{\#Adv.T} &
\shortstack{Clean} &
\shortstack{Att} &
\shortstack{Att+Dec} &
\shortstack{Att+Def} &
\shortstack{Att+Dec+Def} \\

\midrule


\multirow{2}{*}{\textbf{DeiT}}
& 1
& 82.32
& \acc{14.75}{1.11}
& \acc{68.55}{1.45}
& \acc{49.90}{1.56}
& \acc{58.40}{1.54} \\

& 4
& -
& \acc{0.00}{0.00}
& \acc{56.15}{1.55}
& \acc{42.68}{1.55}
& \acc{46.78}{1.56} \\

\midrule


\multirow{2}{*}{\textbf{ViT-B}}
& 1
& 85.06
& \acc{25.78}{1.37}
& \acc{66.60}{1.47}
& \acc{52.34}{1.56}
& \acc{54.39}{1.56} \\

& 4
& -
& \acc{0.00}{0.00}
& \acc{41.70}{1.54}
& \acc{41.50}{1.54}
& \acc{33.50}{1.48} \\

\midrule


\multirow{2}{*}{\textbf{ViT-S}}
& 1
& 81.64
& \acc{11.04}{0.98}
& \acc{54.10}{1.56}
& \acc{50.49}{1.56}
& \acc{44.04}{1.55} \\

& 4
& -
& \acc{0.00}{0.00}
& \acc{25.68}{1.37}
& \acc{26.27}{1.38}
& \acc{14.36}{1.10} \\


\midrule
\multicolumn{7}{c}{%
    Adversarial Patch~\cite{brown2017adversarial}
} \\
\midrule


\multirow{2}{*}{\textbf{DeiT}}
& 1
& 82.32
& \acc{18.16}{1.21}
& \acc{69.04}{1.45}
& \acc{46.58}{1.56}
& \acc{48.83}{1.56} \\

& 4
& -
& \acc{4.49}{0.65}
& \acc{57.23}{1.55}
& \acc{76.86}{1.32}
& \acc{64.06}{1.50} \\

\midrule


\multirow{2}{*}{\textbf{ViT-B}}
& 1
& 85.06
& \acc{18.95}{1.23}
& \acc{59.96}{1.53}
& \acc{65.72}{1.48}
& \acc{58.20}{1.54} \\

& 4
& -
& \acc{0.20}{0.14}
& \acc{41.70}{1.54}
& \acc{83.40}{1.16}
& \acc{80.96}{1.23} \\

\midrule


\multirow{2}{*}{\textbf{ViT-S}}
& 1
& 81.64
& \acc{8.30}{0.86}
& \acc{55.86}{1.55}
& \acc{47.95}{1.56}
& \acc{37.99}{1.52} \\

& 4
& -
& \acc{0.20}{0.14}
& \acc{30.66}{1.44}
& \acc{77.73}{1.30}
& \acc{63.77}{1.50} \\

\bottomrule
\end{tabular}%
}
\caption{\small{
\rev{
Results of combining decoys with the RSA defense under the PatchFool (top) and Adversarial Patch (bottom). Each model is evaluated with $1$ and $4$ adversarial tokens (patches), fixing the number of decoys to $4$. The defense employs a window of \(2\times2\).
}
}}
\label{tab:rsa_results}
\end{table}

\subsection{Position analysis}
\label{app:position_analysis}
Results in Tab.~\ref{tab:position_results} show the difference between applying decoy tokens randomly, as done in the main experiments, and placing them in the top-left corner of the image. This analysis aims to assess whether a fixed position, especially in a corner of the image where positional encodings may play a different role in the attention process, can affect the quality of the decoy.
Experiments here are carried out on a reduced subset of $128$ samples.

As shown in the results, when using a small number of decoys, e.g., one decoy token, placing them in an extreme corner can reduce their ability to control attention compared with random placement. However, this effect becomes less pronounced as the number of decoys increases. With four decoy tokens, the difference in attack success rate becomes negligible. 
On the contrary, the top-$k$ analysis shows slightly different results. Tab.~\ref{tab:topk_positionablation} shows that, even in a corner position, fixed decoy patches are capable of maximizing attention in their area. This is likely due to the fact that fixing the position also arranges the patches into a compact square rather than leaving them sparse, and as the only attention attractors, since here we do not have the underlying attack, they probably work better when grouped than when separated, showing slightly different behaviors in the presence or absence of adversary-attacked regions.

This suggests that future work could further investigate improved decoy placement strategies, possibly considering an attention pre-analysis similar to PatchFool, where the attention of the clean image is first analyzed to determine where to place the attack tokens. We hope to further explore this direction in future work, including the possibility of optimizing the decoy jointly with the attack, rather than relying on the sequential but versatile approach adopted in the main experiments.

\begin{table}[t]
\centering
\scriptsize
\setlength{\tabcolsep}{4pt}
\renewcommand{\arraystretch}{1.15}

\newcommand{\acc}[2]{%
    \ensuremath{#1{\scriptscriptstyle\,\pm\,#2}}%
}

\newcommand{\rowstrut}{\rule{0pt}{2.5ex}}

\newcommand{\ResultRow}[3]{%
    #1
        & \rowstrut #2
        & #3 \\
}

\resizebox{0.65\columnwidth}{!}{%
\begin{tabular}{@{}ccc@{}}
\toprule
\shortstack{\# Decoy tokens} &
\shortstack{Random position $\downarrow$} &
\shortstack{Fixed position $\downarrow$} \\
\midrule

\multicolumn{3}{c}{%
    \textbf{DeiT-B} \quad ARMRO $(\tau=1.2)$
} \\
\midrule

\ResultRow{1}
{\acc{\textcolor{blue!65!black}{54.69}}{4.42}}
{\acc{\textcolor{red!70!black}{59.38}}{4.36}}

\addlinespace[1pt]
\cmidrule(lr){1-3}

\ResultRow{2}
{\acc{\textcolor{blue!65!black}{45.31}}{ 4.42}}
{\acc{\textcolor{red!70!black}{47.66}}{4.43}}

\addlinespace[1pt]
\cmidrule(lr){1-3}

\ResultRow{4}
{\acc{\textcolor{blue!65!black}{27.34}}{3.96}}
{\acc{\textcolor{red!70!black}{27.34}}{3.96}}

\midrule

\multicolumn{3}{c}{%
    \textbf{ViT-S} \quad ARMRO $(\tau=1.4)$
} \\
\midrule

\ResultRow{1}
{\acc{\textcolor{blue!65!black}{31.25}}{ 4.11}}
{\acc{\textcolor{red!70!black}{42.97}}{4.39}}

\addlinespace[1pt]
\cmidrule(lr){1-3}

\ResultRow{2}
{\acc{\textcolor{blue!65!black}{18.75}}{3.46}}
{\acc{\textcolor{red!70!black}{27.34}}{3.96}}

\addlinespace[1pt]
\cmidrule(lr){1-3}

\ResultRow{4}
{\acc{\textcolor{blue!65!black}{15.62}}{3.22}}
{\acc{\textcolor{red!70!black}{15.62}}{3.22}}

\bottomrule
\end{tabular}%
}
\caption{\small{
Effect of decoy placement under ARMRO when using four mask-aware adversarial patches placed at the center of the image. We compare random decoy placement with fixed placement at the first image-token position, corresponding to the top-left corner of the image.}}
\label{tab:position_results}
\end{table}

\begin{table}[t]
\centering
\footnotesize
\setlength{\tabcolsep}{3pt}
\resizebox{\columnwidth}{!}{%
\begin{tabular}{c|cccccccccccc}
\hline
Model & L0 & L1 & L2 & L3 & L4 & L5 & L6 & L7 & L8 & L9 & L10 & L11 \\
\hline

DeiT rand pos & 89 & 42 & 74 & 100 & 100 & 100 & 98 & 96 & 86 & 91 & 96 & 100 \\
DeiT fix pos & 98 & 57 & 93 & 100 & 100 & 100 & 100 & 98 & 99 & 99 & 98 & 100 \\
\hline

ViT-B rand pos & 29 & 85 & 89 & 100 & 96 & 85 & 49 & 52 & 17 & 18 & 35 & 99 \\
ViT-B fix pos & 68 & 93 & 94 & 99 & 98 & 90 & 36 & 45 & 19 & 20 & 25 & 100 \\
\hline

ViT-S rand pos & 88 & 97 & 93 & 100 & 100 & 100 & 96 & 43 & 42 & 46 & 64 & 99 \\
ViT-S fix pos & 92 & 100 & 91 & 100 & 100 & 100 & 97 & 76 & 77 & 81 & 92 & 100 \\
\hline
\end{tabular}
}
\caption{\small{Top-k overlap (\%) per layer for DeiT, ViT-B, and ViT-S maintaining the random positioning (rand pos) or by fixing the position to the left top corner (fix pos) of $4$ decoy tokens.}}
\label{tab:topk_positionablation}
\end{table}

\subsection{Adaptive attack and patch-budget analysis.}
\label{appx:adaptive_attack}
The adaptive attack optimization used in Sec.~\ref{sec:budget_analysis} and in the following experiments follows a similar idea to the optimization presented in App.~\ref{app:attention_analysis}, where we study the effect of reducing attention while preserving adversarial effectiveness. In particular, we consider all network layers, since defenses such as ARMRO are applied across all layers. We set $\lambda$ to $0.01$, which provides the best empirical compromise between reducing attention and avoiding an excessive degradation of adversarial effectiveness.

\begin{figure}[ht]
    \centering
    \begin{subfigure}[t]{\textwidth}
    \begin{subfigure}[t]{0.24\textwidth}
    \includegraphics[width=\linewidth]{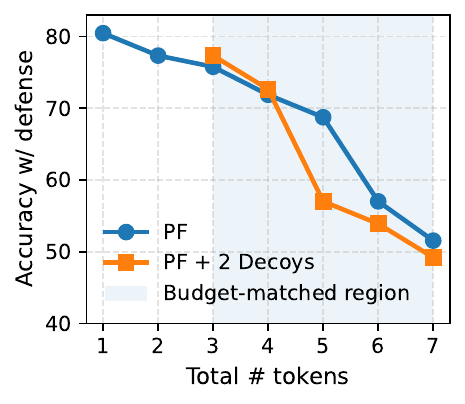}
    \end{subfigure}
    \begin{subfigure}[t]{0.24\textwidth}
    \includegraphics[width=\linewidth]{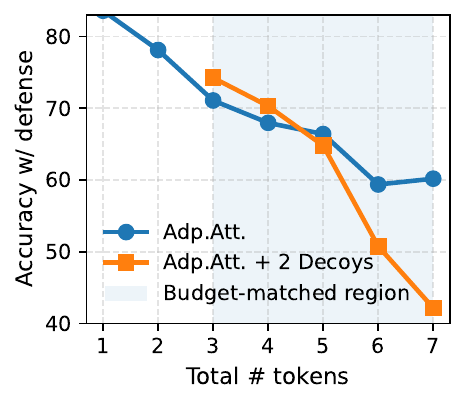}
    \end{subfigure}
    \end{subfigure}
    \begin{subfigure}[t]{\textwidth}
    \begin{subfigure}[t]{0.24\textwidth}
    \includegraphics[width=\linewidth]{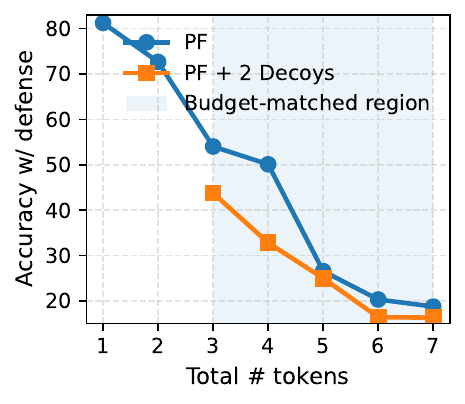}
    \end{subfigure}
    \begin{subfigure}[t]{0.24\textwidth}
    \includegraphics[width=\linewidth]{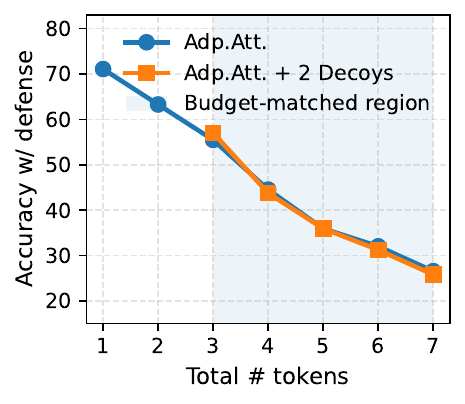}
    \end{subfigure}
    \end{subfigure}
    \vspace{-1em}
    \caption{\small{Patch-budget comparison on ViT-B (top) and ViT-S (bottom), between using all tokens for the attack and reserving $2$ tokens for decoys.}}
    \label{fig:budget_analysis_suppl}
\end{figure}

Fig.~\ref{fig:budget_analysis_suppl} reports an additional patch-budget analysis, computed on a reduced subset of $128$ samples, comparing the use of all available tokens for the attack with the proposed setting, where part of the token budget is reserved for decoys. This analysis is important because decoy tokens increase the total number of modified tokens, and therefore the comparison should also account for attacks that use the same overall budget entirely for adversarial optimization. Across both ViT-B and ViT-S, and for both adaptive adversarial patches and PatchFool, increasing the number of attack tokens generally improves the attack effectiveness, as expected. However, the results show that reserving part of the budget for decoys remains competitive, and in several cases more effective, than allocating the entire budget to adversarial tokens only. These results indicate that the advantage of decoys is not merely due to increasing the perturbation budget, but rather to their ability to redirect the masking defense away from the truly adversarial regions. 

\section{Additional illustration}
Finally, Fig.~\ref{fig:additional_illustration} reports an illustration of the attention activation across layers for different versions of the image: clean, attacked, attack with defense, attack with decoy, and attack with decoy and defense.

\begin{sidewaysfigure*}[ht]
    \centering
    \includegraphics[width=\linewidth]{new_images/imagenet_heatmap_appendix_new.png}
    \vspace{-1em}
    \caption{\small{Qualitative illustration of the attention activations across ViT layers for different image configurations. Each row corresponds to a transformer layer, while columns show the clean image, the attacked image, the attacked image after defense, the attack with decoys, and the attack with decoys after defense.}}
    \label{fig:additional_illustration}
\end{sidewaysfigure*}


\end{document}